\pdfoutput=1
\documentclass{article}

\usepackage[preprint]{neurips_2024}

\usepackage[utf8]{inputenc} %
\usepackage[T1]{fontenc}    %
\usepackage{hyperref}       %
\usepackage{url, eucal}            %
\usepackage{booktabs}       %
\usepackage{amsfonts}       %
\usepackage{nicefrac}       %
\usepackage{microtype}      %

\usepackage[table]{xcolor}
\usepackage{amsmath}
\usepackage{caption, subcaption, multirow, overpic, textpos}
\usepackage{booktabs}
\usepackage{xspace}
\usepackage{balance}
\usepackage{numprint}
\usepackage{sidecap}
\usepackage{float}
\usepackage{algorithm}
\usepackage{algorithmic}
\usepackage{tcolorbox}
\usepackage{adjustbox}
\usepackage{tabularx}
\usepackage{makecell}
\usepackage{graphicx}

\usepackage{natbib}
\setcitestyle{square,comma,numbers}

\usepackage[capitalize]{cleveref}
\crefname{section}{Sec.}{Secs.}
\Crefname{section}{Section}{Sections}
\Crefname{table}{Table}{Tables}
\crefname{table}{Table}{Tables}

\newcommand{\etal}{\textit{et al.}}
\definecolor{mycolor}{gray}{.69}

\usepackage{hyperref}
\hypersetup{
    colorlinks=true,
    breaklinks=true,
    urlcolor=blue,
    linkcolor=blue,
    bookmarksopen=false,
    filecolor=black,
    citecolor=mycolor,
    linkbordercolor=blue
}

\title{GeoBench: Benchmarking and Analyzing \\Monocular Geometry Estimation Models}

\author{%
  Yongtao Ge\textsuperscript{1,2}, \quad Guangkai Xu\textsuperscript{1}, \quad Zhiyue Zhao\textsuperscript{1}, \quad Libo Sun\textsuperscript{2}, \\ 
  \bf
  Zheng Huang\textsuperscript{1}, \quad Yanlong Sun\textsuperscript{3}, \quad Hao Chen\textsuperscript{1}, \quad Chunhua Shen\textsuperscript{1} \\[0.25cm]    
  \textsuperscript{1} Zhejiang University \quad \textsuperscript{2} The University of Adelaide \quad \textsuperscript{3}  Tsinghua University \\[0.25cm]
  Code \& Benchmark: \url{https://github.com/aim-uofa/GeoBench}
}

\begin{document}

\maketitle
\newcommand{\predTheta}{\hat{\mathbf{\Theta}}}
\newcommand{\predShape}{\hat{\mathbf{\beta}}}
\newcommand{\predPose}{\hat{\mathbf{\theta}}}
	
\newcommand{\gtTheta}{\mathbf{\Theta}}
\newcommand{\gtShape}{\mathbf{\beta}}
\newcommand{\gtPose}{\mathbf{\theta}}

\newcommand{\motionDisc}{\mathcal{D}_M}	
\newcommand{\generator}{\mathcal{G}}
\newcommand{\smpl}{\mathcal{M}}	

\newcommand{\focalx}{f_x}
\newcommand{\focaly}{f_y}	
\newcommand{\centerx}{o_x}
\newcommand{\centery}{o_y}
\newcommand{\campos}{C}
\newcommand{\camtransl}{t^c}
\newcommand{\camrot}{R^c}
\newcommand{\imgwidth}{w}
\newcommand{\imgheight}{h}
\newcommand{\cropwidth}{w_{bbox}}
\newcommand{\cropheight}{h_{bbox}}
\newcommand{\cropcenterx}{c_x}
\newcommand{\cropcentery}{c_y}
\newcommand{\predjoints}[1]{\hat{\mathcal{J}}_{\mathit{#1D}}}
\newcommand{\gtjoints}[1]{\mathcal{J}_{\mathit{#1D}}}

\newcommand{\bodytransl}{t^b}
\newcommand{\bodyori}{R^b}

\newcommand{\pampjpe}{PA-MPJPE\xspace}
\newcommand{\wmpjpe}{W-MPJPE\xspace}
\newcommand{\cmpjpe}{C-MPJPE\xspace}
\newcommand{\mpjpe}{MPJPE\xspace}
\newcommand{\wpve}{W-PVE\xspace}
\newcommand{\cpve}{C-PVE\xspace}
\newcommand{\pve}{PVE\xspace}

\newcommand{\iwcam}{IWP-cam\xspace}
\newcommand{\camcalib}{CamCalib\xspace}
\newcommand{\pitch}{\alpha}
\newcommand{\roll}{\phi}
\newcommand{\yaw}{\psi}
\newcommand{\vfov}{\upsilon}
\newcommand{\softltwo}{Softargmax-$\mathcal{L}_{2}$\xspace}
\newcommand{\softbiasedltwo}{Softargmax-biased-$\mathcal{L}_{2}$\xspace}
\newcommand{\ltwo}{$\mathcal{L}_2$\xspace}
\newcommand{\smplify}{SMPLify-X-cam\xspace}

\newcommand{\methodname}{SPEC\xspace} %
\newcommand{\figref}[1]{Fig.~\ref{#1}}

\newcommand{\mpi}{\texttt{MPI-INF-3DHP}\xspace}
\newcommand{\mpii}{\texttt{MPII}\xspace}
\newcommand{\lspet}{\texttt{LSPET}\xspace}
\newcommand{\hthreesixm}{\texttt{Human3.6M}\xspace}
\newcommand{\threedpw}{\texttt{3DPW}\xspace}
\newcommand{\threedpwocc}{\texttt{3DPW-OCC}\xspace}
\newcommand{\coco}{\texttt{COCO}\xspace}
\newcommand{\cocoeft}{\texttt{COCO-EFT}\xspace}
\newcommand{\ooh}{\texttt{3DOH}\xspace}

\newcommand{\agoracam}{\methodname-SYN\xspace}
\newcommand{\mtpcam}{\methodname-MTP\xspace}

\newcommand{\smplifyxc}{SMPLify-XC\xspace}

\newcommand{\supmat}{Sup.~Mat.\xspace}

\newcommand{\real}{\mathbb{R}}

\def\etal{\emph{et al}.}
\def\eg{\emph{e.g}.}
\def\ie{\emph{i.e}.}
\def\vs{\emph{vs}.\xspace}

\newcommand{\bs}[1]{\boldsymbol{#1}}
\newcommand{\vecb}[1]{{#1}}

\definecolor{myyellow}{rgb}{0.8,0.8,0}
\definecolor{mygreen}{rgb}{0,0.8,0}
\definecolor{myred}{rgb}{0.8,0,0}

\newcommand{\wcon}{W_{\scalebox{0.7}{\textnormal{cn}}}}
\newcommand{\del}{\textnormal{zr}}

\begin{abstract}

Recent advances in discriminative and generative pretraining have yielded geometry estimation models with strong generalization capabilities. While discriminative monocular geometry estimation methods rely on large-scale fine-tuning data to achieve zero-shot generalization,
several generative-based paradigms show the potential of achieving impressive generalization performance on unseen scenes by leveraging pre-trained diffusion models and fine-tuning on even a small scale of synthetic training data.
Frustratingly, these models are trained with different recipes on different datasets, making it hard to find out the critical factors that determine the evaluation performance.
Besides, the current widely used geometry evaluation benchmarks have two main drawbacks that may prevent the development of the field, \ie, \emph{limited scene diversity} and \emph{unfavorable label quality}.
To resolve the above issues, (1) we \emph{\textbf{build fair and strong baselines in a unified codebase}} for evaluating and analyzing the state-of-the-art (SOTA) geometry estimation models in terms of both different finetuning paradigms and training recipes;
(2) we \emph{\textbf{evaluate monocular geometry estimators on more challenging benchmarks}} for geometry estimation task with
diverse scenes and high-quality annotations.

Our results reveal that pre-trained using large data, discriminative models such as DINOv2, can outperform generative counterparts with  \textit{\textbf{a small amount of high-quality synthetic training data}} under the same training configuration, which suggests that fine-tuning data quality is a more important factor than the data scale and model architecture.
Our observation also raises a question: if simply fine-tuning a general vision model such as DINOv2 using a small amount of synthetic depth data produces SOTA results, \textit{do we really need complex models, \textit{e.g.},  Marigold \cite{marigold} and DepthFM \cite{depthfm} for depth estimation?} 
We believe that this work can propel advancements in geometry estimation tasks 
and a wide range of other downstream vision tasks.


\end{abstract}

\section{Introduction}
\label{sec:intro}

Monocular depth and surface normal estimation, also referred to as ``monocular geometry estimation'', poses a fundamental yet intricate challenge of inferring distance and surface orientation from a single image. Its significance is underscored by its broad utility across various downstream tasks, including object detection~\cite{huang2022monodtr, wang2020task, ding2020learning}, visual navigation~\cite{tateno2017cnn, yang2020d3vo, sun2022improving, yang2018deep}, novel view synthesis~\cite{deng2022depth, roessle2022dense}, controllable image generation~\cite{2023_ControlNet, esser2023structure, zhao2024uni}, and 3D scene reconstruction \cite{sun2021neuralrecon, denninger20203d}. The importance of this task has led to a significant body of research, resulting in numerous models~\cite{birkl2023midas, depthanything, yin2023metric, hu2024metric3dv2, marigold} over the past decade.

Although a large number of monocular geometry estimation models exist, they can be divided into two paradigms, \ie, discriminative-based and generative-based.
Most contemporary discriminative monocular geometry estimation models leverage the pre-train priors from fully-supervised image classification backbones, \eg, ConvNeXt~\cite{woo2023convnext}, EfficientNet~\cite{tan2019efficientnet} and ViT~\cite{vit}, or self-supervised backbones. \eg, DINOv2~\cite{dinov2}, previous best discriminative depth estimation models scale-up their performance with pre-trained DINOv2 backbone and \textit{a large scale of fine-tuning data}. Generative geometry estimation 
models~\cite{marigold,geowizard,genpercept,depthfm,lee2024dmp} unleash the power of pre-trained text-to-image diffusion models, \eg, Stable Diffusion (SD)~\cite{ldm}.
Several generative geometry estimation models~\cite{marigold, geowizard, genpercept, depthfm} show strong generation capability with even \textit{a small-scale high-quality synthetic fine-tuning data}. %

However, none of the previous works have systematically investigated the performance of these geometry estimation methods with fair and faithful comparison. 
The reason is twofold. Firstly, \textit{the different selections of datasets and training configurations hinder the fair evaluations of the newly designed methodologies.} \textbf{(1)} The performance distinction for different generative-based finetuning paradigms is unclear. It is hard to evaluate whether the actual improvement is from the algorithmic perspective or the data perspective since they are trained on different datasets and different training configurations. \textbf{(2)} The performance distinction between discriminative and generative geometry estimation models when trained on the same scale and quality of data also remains unclear.
Secondly, \textit{existing popular geometry estimation benchmarks may not reveal the real performance of the models.} NYUv2~\cite{DATA_2012_NYUv2} and  ScanNet~\cite{DATA_2017_ScanNet} are still popular in the evaluation of indoor monocular depth estimation. However, they are collected by an older Kinect-v1 system with noisy depth measurements and noisy imaging for RGB patterns, with only $640\times480$ resolution.
DIODE~\cite{DATA_2019_DIODE} and ETH3D~\cite{DATA_2017_ETH3D} collect both outdoor and indoor scenes with high-quality data while with low diversity scenes for evaluation.
KITTI~\cite{DATA_2012_KITTI} collects depth maps from the LIDAR sensor and focuses on outdoor driving scenes.
For surface normal evaluation, NYUv2~\cite{DATA_2012_NYUv2}, ScanNet~\cite{DATA_2017_ScanNet}, iBims-1~\cite{DATA_2018_iBims}, Sintel~\cite{DATA_2012_SINTEL} and Virtual KITTI~\cite{DATA_2016_VKITTI} are widely used by generating surface normal maps from the ground truth depth maps. However, the depth noises in NYUv2~\cite{DATA_2012_NYUv2}, ScanNet~\cite{DATA_2017_ScanNet} and iBims-1~\cite{DATA_2018_iBims} yield unsatisfactory surface normal ground truth. The limited scene diversity of synthetic datasets, \ie, Sintel~\cite{DATA_2012_SINTEL} and Virtual KITTI~\cite{DATA_2016_VKITTI}, cannot evaluate the robustness of the surface normal estimation model for in-the-wild geometry reconstruction. Overall, the existing geometry benchmarks are hindered by two main issues: ground-truth quality and scene diversity. This lack of fair and comprehensive benchmarks can significantly impede the development of geometry estimation research. 

To address the aforementioned problems, we perform a comprehensive geometry estimation benchmarking study from two perspectives:

\begin{enumerate}
    \item


\textbf{Training strategy.} We reimplement a 
comprehensive list of 
SOTA algorithms in a unified codebase, including  Marigold~\cite{marigold}, Geowizard~\cite{geowizard}, GenPercept~\cite{genpercept}, DepthFM~\cite{depthfm}, DMP~\cite{lee2024dmp}, Depth-Anything~\cite{depthanything} and DSINE~\cite{bae2024dsine}. As such, we can fairly evaluate their performance under the same training configuration, and figure out whether the performance improvement is coming from the model architecture or coming from the high-quality training data. Previous generative geometry models are all based on Stable Diffusion 2.1~\cite{ldm} with limited trained data (compared to the data scale of discriminative models), we further explore the potential of generative geometry models by conducting data scale-up and model scale-up ablations.  


\item 
\textbf{More benchmark datasets.} Apart from traditional geometry evaluation benchmarks, we build more diverse scenes with high-quality labels for geometry evaluation. For depth estimation, we introduce three extra benchmark datasets, InSpaceType~\cite{wu2023inspacetype}, MatrixCity~\cite{li2023matrixcity}, and Infinigen~\cite{infinigen}. InSpaceType is an 
indoor depth evaluation benchmark, which contains 12 scenes, 1260 images, and $2208\times1242$ resolution. It is a good complement for indoor benchmarks like NYUv2 and ScanNet. MatrixCity is a rendered dataset with real city-scale scenes, we select 808 street images and 403 aerial images for evaluation. It is suitable for evaluating driving and city scenes. Infinigen is also a high-quality rendered dataset, which contains diverse nature scenes. We use it to verify the generalization capability of depth estimation foundation models in wild scenes. For surface normal estimation, we expand existing benchmark datasets with more high quality and diverse datasets, \eg, indoor MuSHRoom dataset~\cite{DATA_2024_MUSHROOM}, outdoor Tank and Temples (T\&T) dataset~\cite{DATA_2017_Tanks_and_Temples}\footnote{The surface normal annotation of MushRoom and T\&T are obtained from Gaustudio~\cite{ye2024gaustudio}}, and wild Infinigen~\cite{infinigen} dataset. 
\end{enumerate}

With the unified codebase, training data, and comprehensive benchmark datasets, we surprisingly find that: 
\begin{enumerate}
    \item 

Without bells and whistles, 
the  
depth estimation model pre-trained on the DINOv2~\cite{dinov2} backbone, followed by a simple DPT head~\cite{MDE_2021_DPT}, can already achieve great generalization performance by using only a small amount of high-quality synthetic datasets ($77$K training samples). 

\textit{With the same training configuration, the discriminative model outperforms the recent generative counterpart, Marigold~\cite{marigold}.}
Besides, the result 
is a sharp contrast to previous discriminative models using large data to achieve generalization capability, \eg, Metric3Dv2~\cite{hu2024metric3dv2}, which focuses on collecting more diverse training datasets ($16$M training samples), and Depth-Anything~\cite{depthanything}, which focuses on scaling up model performance with large-scale pseudo labels ($63.5$M training samples). \textit{Based on the result, we argue that training data quality may be more important than the data scale.}

\item 
Generative-based geometry estimation models can generate high-resolution and detailed depth and surface normal maps, which is an advantage over the existing discriminative depth estimation models. We mainly attribute this to high-quality synthetic fine-tuning data and the design of the VAE~\cite{vae} decoder, which can decode compressed depth latent to the same resolution depth map as the input image. 

\item 
For surface normal estimation, the discriminative model DSINE~\cite{bae2024dsine} outperforms generative-based fine-tuning protocols with the same training configuration, 
suggesting that, apart from large-scale pre-training, inductive bias~\cite{bae2024dsine} can also be an important factor in providing 
rich 
information for surface normal estimation tasks. 

\end{enumerate}

We hope that our benchmarking results 
can 
pave the way for designing more powerful geometry estimation algorithms and developing more high-quality geometry estimation training datasets in the future.

\section{Preliminaries}
\textbf{Task definition.}
Given an input image $x\in \mathbb{R}^{H \times W \times 3}$, the goal of monocular geometry estimation is to predict the depth map $d\in \mathbb{R}^{H \times W}$, which can be affine-invariant or metric depth, and surface orientation, which can be represented as either a unit vector $\mathbf{n}\in S^2$, or a 3D \textit{axis-angle} $\mathbf{R} \in SO(3)$.

\textbf{Discriminative geometry estimation models.}
With the widespread application of deep learning~\cite{lecun2015deep}, learning-based methods have demonstrated their ability to estimate geometric information from monocular images~\cite{eigen2014depth, godard2019digging, Ranftl2022, depthanything}. Early works primarily relied on discriminative models using either supervised or unsupervised methods. Eigen et al.~\cite{eigen2014depth} proposed the first learning-based method for monocular depth estimation, employing two deep network stacks and using ground truth depth for supervision. Zhou et al. proposed an early unsupervised framework, SfMLearner~\cite{zhou2017unsupervised}, in which camera pose and monocular depth are learned together. With the availability of large amounts of data, recent methods~\cite{Ranftl2022, depthanything, yin2023metric, hu2024metric3dv2} have shown a trend toward using large-scale datasets to develop robust geometry estimation models that generalize well to diverse environments. For instance, Ranftl et al.~\cite{Ranftl2022} introduced a method that demonstrates strong zero-shot testing ability by utilizing mixed training datasets. Yang et al.~\cite{depthanything} further improved zero-shot monocular depth estimation performance by proposing Depth-Anything, which leverages large-scale data to achieve strong generalization ability. Meanwhile, Yin et al.~\cite{yin2023metric, hu2024metric3dv2} proposed Metric3D, which can output accurate metric depth by training models on 11 public RGB-D datasets. Apart from depth estimation, advancements in surface normal information have also been achieved through the use of discriminative models. Surface normal information can not only be calculated directly from depth maps but can also be independently obtained through surface normal estimation techniques~\cite{wang2015designing, ladicky2014discriminatively, lenssen2020deep, bae2024dsine}. For example, Bae et al.~\cite{bae2024dsine} proposed a method that demonstrates strong generalization capabilities and produces high-quality surface normal predictions by investigating inductive biases. Overall, the use of discriminative models for both depth and surface normal estimation has shown its significance in improving performance, thereby broadening the applications of monocular geometry estimation.

\textbf{Generative geometry estimation Models.}
Given the impressive results of recent generative models~\cite{ldm} in image generation tasks, many studies have endeavored to incorporate generative-based pipelines into geometry estimation. Ji et al.~\cite{ji2023ddp} proposed a method to extend the denoising diffusion process into the modern perception pipeline, which can be generalized to most dense prediction tasks, such as depth estimation. Saxena et al.~\cite{saxena2024surprising} formulated optical flow and monocular depth estimation as image-to-image translation using generative diffusion models, without specialized loss functions and model architectures. Zhao et al.~\cite{zhao2023unleashing} proposed VPD, a framework that exploits the semantic information of a pre-trained text-to-image diffusion model in visual perception tasks. Ke et al.~\cite{marigold} introduced a method for affine-invariant monocular depth estimation, where the depth information is derived from retained rich stable diffusion priors. Fu et al.~\cite{geowizard} proposed a foundation model for jointly estimating depth and surface normal from monocular images, which not only achieves surprisingly robust generalization on various types of real or synthetic images but also faithfully captures intricate geometric details.
In summary, recent generative-based methods have provided new solutions and demonstrated their applications for depth estimation.

\textbf{Geometric evalutaion metrics.}
We use widely adopted evaluation metrics for assessing the performance of depth and surface normal estimation. Specifically, for the depth estimation task, we use mean absolute relative error (AbsRel) and accuracy under thresholds ($\delta_{i} < 1.25^{i}, i=1, 2, 3$) for accuracy comparisons. These evaluation metrics for depth estimation are calculated as follows:
\textbf{(1)} mean absolute relative error (AbsRel): $\frac{1}{n}\sum_{i=1}^{n} \frac{\left | z_{i}-z_{i}^{\ast} \right |}{z_{i}^{\ast}}$;
\textbf{(2)} the accuracy under threshold ($\delta_{i} < 1.25^{i}, i=1, 2, 3$): $ \%$ of $z_{i}$  s.t. \rm{max}$\left (\frac{z_{i}}{z_{i}^{\ast}}, \frac{z_{i}^{\ast}}{z_{i}} \right ) < 1.25^{i}$;
where $z_i$ is the ground truth depth and $z_{i}^{\ast}$ represents the predicted depth. For surface normal estimation, we calculate the angular error for the pixels with ground truth and report both the median and mean values (lower is better). In addition, we measure the percentage of pixels with an error below $t \in [5.0^\circ, 11.25^\circ, 30.0^\circ]$ (higher is better). Please refer to~\cite{bae2024dsine} for calculation details.

\section{Benchmarking Depth Estimation Foundation Models}

\subsection{A Brief Overview of SOTA Methods}
\begin{table}[tb!]
\footnotesize
\setlength\tabcolsep{1.5pt}
\renewcommand{\arraystretch}{1.0} %

\centering

\caption{Quantitative comparison on 5 zero-shot affine-invariant depth benchmarks \textbf{with author released weights}. We mark the best results in bold and the second best underlined. Discriminative methods are colored in \colorbox{blue!20}{blue} while generative ones in \colorbox{green!20}{green}.}  

\resizebox{\textwidth}{!} 
{
\begin{tabular}{c|c|c|cccccccccc}
\toprule
\multirow{2}{*}{Method}  & \multirow{2}{*}{\shortstack{Train\\Samples}} & \multirow{2}{*}{Year} &  \multicolumn{2}{c}{ NYUv2~\cite{DATA_2012_NYUv2} } & \multicolumn{2}{c}{ KITTI~\cite{DATA_2012_KITTI} } & \multicolumn{2}{c}{ ETH3D~\cite{DATA_2017_ETH3D} } & \multicolumn{2}{c}{ ScanNet~\cite{DATA_2017_ScanNet} } & \multicolumn{2}{c}{ DIODE~\cite{DATA_2019_DIODE}} \\
\cline{4-13}
& & & AbsRel $\downarrow$ & $\delta 1 \uparrow$ & AbsRel $\downarrow$ & $\delta 1 \uparrow$ & AbsRel $\downarrow$ & $\delta 1 \uparrow$ & AbsRel $\downarrow$ & $\delta 1 \uparrow$ & AbsRel $\downarrow$ & $\delta 1 \uparrow$ \\
\midrule
\rowcolor{blue!10} Metric3Dv2~\cite{hu2024metric3dv2} & 16M & arXiv'24 & \textbf{3.9} & 97.9 & \textbf{5.2} & \textbf{97.9} & \textbf{4.0} & \textbf{98.3} & \textbf{2.3} & \textbf{98.9} & \textbf{14.7} & \textbf{89.2}\\
\midrule
\rowcolor{blue!10} DepthAnything~\cite{depthanything} & 63.5M & CVPR'24 & \underline{4.3} & \textbf{98.0} & \underline{8.0} & \underline{94.6} & \underline{5.8} & \underline{98.4} & \underline{4.3} & \underline{98.1} &\underline{26.1} & 75.9 \\
\midrule
\rowcolor{green!10} Marigold~\cite{marigold} & 74K & CVPR'24 & 5.5 & 96.4 & 9.9 & 91.6 & {6.5} & {96.0} & 6.4 & 95.1 & 30.8 & {77.3}  \\

\midrule
\rowcolor{green!10} GeoWizard~\cite{geowizard} & 280K & arXiv'24 & 5.9 & 95.9 & 12.9 & 85.1 & 7.7 &  94.0 & {6.6} & {95.3} & 32.8 & {75.3} \\
\midrule
\rowcolor{green!10} GenPercept~\cite{genpercept} & 74K & arXiv'24 & 6.3 & 96.0 & 13.3 & 84.1 & 7.2 & 95.5 & 6.6 & 96.0 & 32.3 & 76.0 \\
\midrule
\rowcolor{green!10} DepthFM~\cite{depthfm} & 63K & arXiv'24 & 8.2 & 93.2 & 17.4 & 71.8 & 10.1 & 90.2 & 9.5& 90.3 & 33.4 & 72.9 \\
\bottomrule
\end{tabular}
}
\label{tab: sota_depth_benchmark}
\end{table}

To demonstrate the performance of the SOTA methods, we consider some latest and representative algorithms, \ie, two discriminative models, (Metric3Dv2~\cite{hu2024metric3dv2}, Depth-Anything~\cite{depthanything}), and four generative models (Marigold~\cite{marigold}, DepthFM~\cite{depthfm}, Geowizard~\cite{geowizard} and GenPercept~\cite{genpercept}). We fairly evaluate their performance by using the official released model weights on 5 popular benchmarks, \ie, NYU v2~\cite{DATA_2012_NYUv2}, KITTI~\cite{DATA_2012_KITTI}, ETH3D~\cite{DATA_2017_ETH3D}, ScanNet~\cite{DATA_2017_ScanNet} and DIODE~\cite{DATA_2019_DIODE}, in~\cref{tab: sota_depth_benchmark}. Notably, all the methods do not use these benchmarks as training data. We can easily observe that \textbf{(1)} Metric3Dv2~\cite{hu2024metric3dv2} achieves the best performance on all evaluation datasets, another discriminative-based method, Depth-Anything~\cite{depthanything} achieves the second best performance. Both of them are trained on large-scale datasets, with 16M and 63.5M training data separately. \textbf{(2)} Generative methods can achieve impressive results on these evaluation benchmarks with even a small amount of fine-tuning data.

\begin{figure}[t]
    \centering
    \includegraphics[width=\linewidth]{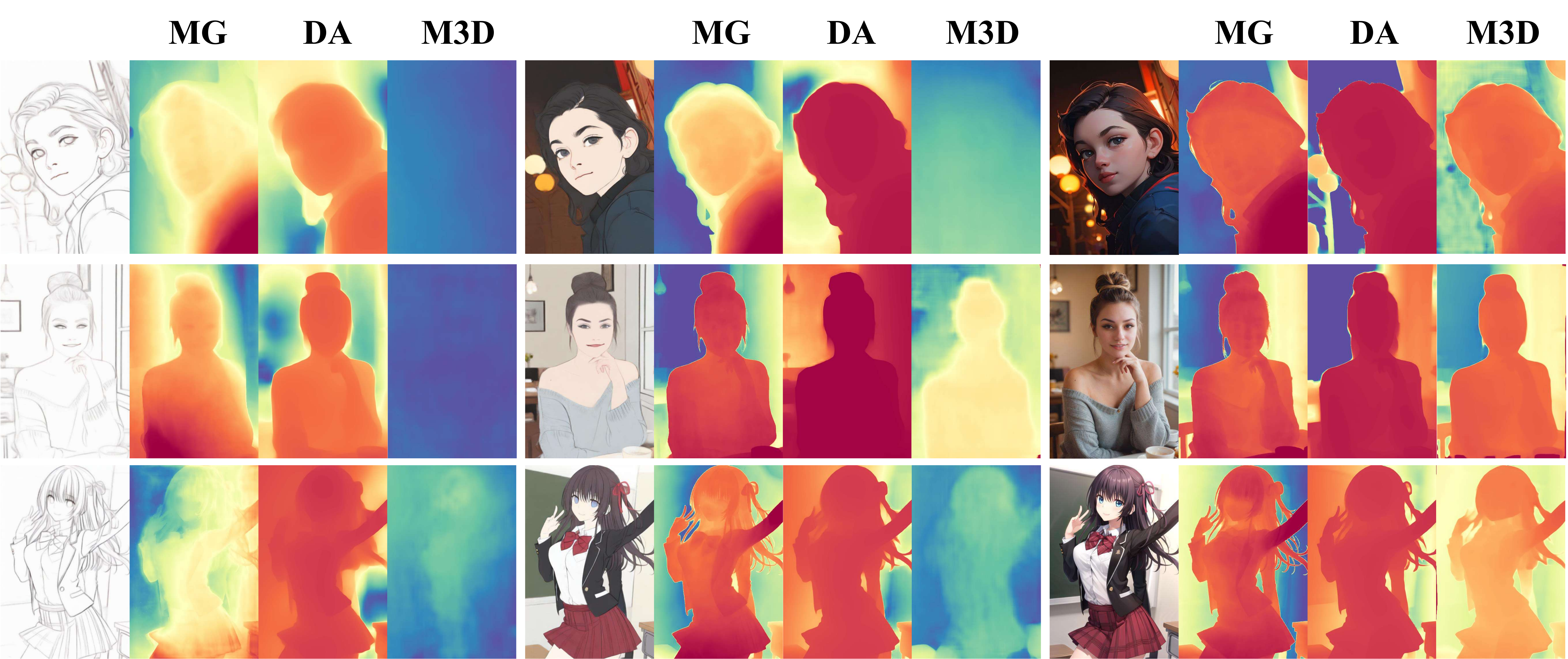}
    \caption{Depth visualization on cartoon images. `MG' indicates Marigold~\cite{marigold}, `DA' indicates Depth-Anything~\cite{depthanything}, `M3D' indicates Metric3Dv2~\cite{hu2024metric3dv2}.}
    \label{fig:cartoon}
\end{figure}

\begin{figure}
    \centering
    \includegraphics[width=0.999\linewidth]{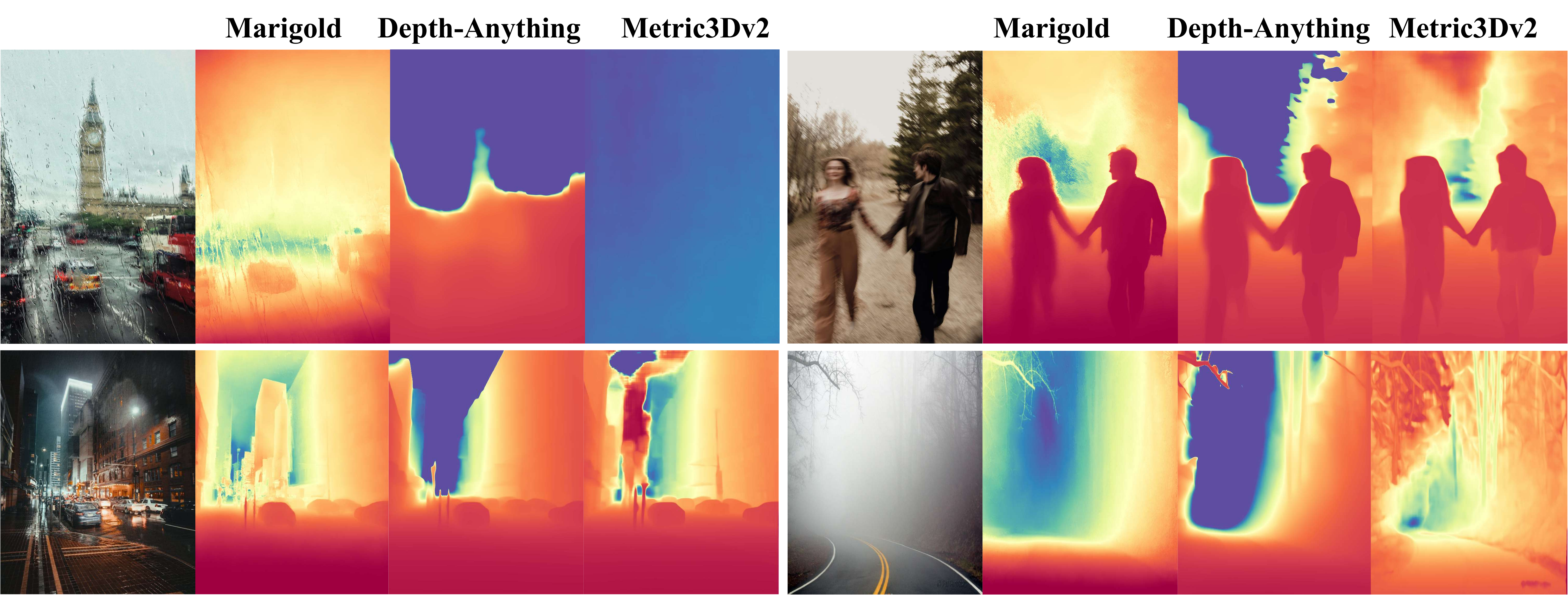}
    \caption{Depth visualization on four challenging scenes, \ie, rainy (top-left), blurry (top-right), dark (bottom-left), and foggy (bottom-right) environments.}
    \label{fig:hard_scenes}
\end{figure}

\begin{figure}
    \centering
    \includegraphics[width=0.999\linewidth]{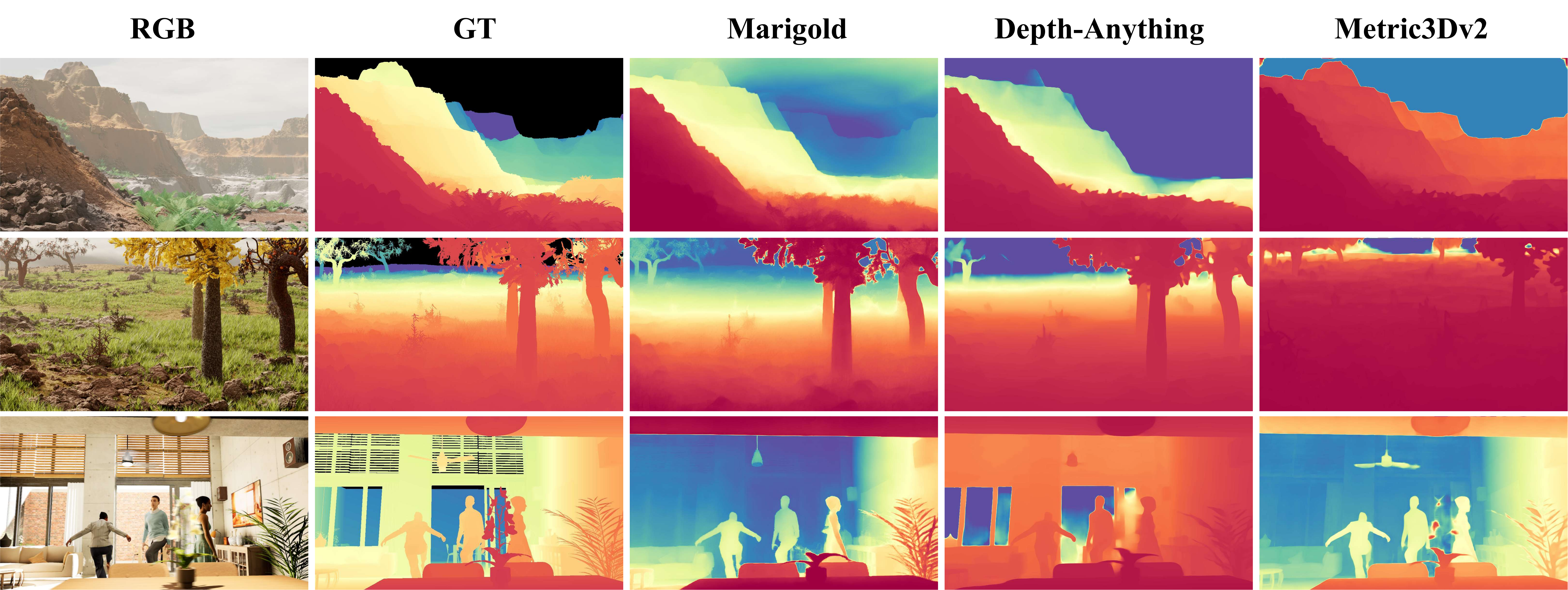}
    \caption{Fine-grained depth estimation comparison. We select two scenes (first two rows) from the Infinigen Dataset~\cite{infinigen} and one scene (last row) from the BEDLAM dataset~\cite{bedlam}.}
    \label{fig:inifinigen_vis}
\end{figure}

In addition to quantitative results, we further test their generalization capability by qualitative visualization in several challenging scenes. ~\cref{fig:cartoon} demonstrates the results of three algorithms on line drawing images (left), color draft images (middle), and photo-realistic images (right). Surprisingly, Metric3D fails on both line draw images and color draft images, while Marigold~\cite{marigold} and Depth-Anything~\cite{depthanything} show some generalization capability on this kind of non-geometrically consistent hand-drawn images. We conjecture that discriminative-based Metric3D does not see cartoon images in the training stage, which leads to poor performance in this scenario. Contrarily, although Marigold~\cite{marigold} also does not see cartoon images in their training set, it leverages the priors stored in the pre-trained Stable Diffusion~\cite{ldm} model. Stable Diffusion~\cite{ldm} model has seen millions of text-cartoon pairs when performing text-to-image generation training.
~\cref{fig:hard_scenes} shows the robustness of existing depth estimation models on challenging scenes like rainy, blurry, dark, and foggy environments. Both Metric3D and Depth-Anything fail on the rainy scene; both Marigold and Metric3D fail to estimate the sky in the second blurry scene. None of the algorithms can handle all environments perfectly.
~\cref{fig:inifinigen_vis} illustrates the depth estimation results on the Infinigen~\cite{infinigen} dataset (first two lines) and BEDLAM~\cite{bedlam} dataset (last line). Infinigen~\cite{infinigen} is a photo-realistic rendered dataset with diverse nature scenes. BEDLAM~\cite{bedlam} is a human-centered high-quality rendered dataset with versatile indoor and outdoor scenes. Mainstream depth evaluation metrics overlook the depth accuracy on the edges of the objects. We use these two datasets to demonstrate the fine-grained depth estimation results since both datasets have high-quality annotations. For measuring the accuracy of depth estimation on edges. We use Canny~\cite{canny1986computational} edge detector
to extract the edge mask from the image and then calculate the traditional depth metrics. As shown in~\cref{tab: fine-grained-depth}, Depth-Anything achieves the highest performance on the Infinigen dataset; Marigold achieves the best AbsRel on the BEDLAM~\cite{bedlam} dataset. 

In a nutshell, discriminative models trained on large data, \ie, Depth-Anything~\cite{depthanything}, 
obtains 
the highest performance in most cases, while generative models fine-tuned on small data, \eg, Marigold~\cite{marigold}, show competitive generalization capability on unseen scenes.

\begin{table}[tb!]
\footnotesize
\setlength\tabcolsep{1.5pt}
\renewcommand{\arraystretch}{1.0} %

\centering

\caption{Benchmarking depth estimation on Infinigen~\cite{infinigen} and BEDLAM~\cite{bedlam} datasets. `Standard' indicates using standard evaluation metrics. `Canny' indicates only evaluating the performance on pixels that belong to canny edges. We mark the best results in bold.} 
\resizebox{1.\textwidth}{!}
{
\begin{tabular}{c|c|cc|cc|cc|cc}
\toprule
\multirow{2}{*}{Method}  & \multirow{2}{*}{\shortstack{Train\\Samples}} &   \multicolumn{2}{c}{ Infinigen-Standard} & 
\multicolumn{2}{c}{ Infinigen-Canny } & 
\multicolumn{2}{c}{ BEDLAM-Standard } & 
\multicolumn{2}{c}{ BEDLAM-Canny}  \\
\cline{3-10}
& & AbsRel $\downarrow$ & $\delta 1 \uparrow$ & AbsRel $\downarrow$ & $\delta 1 \uparrow$ & AbsRel $\downarrow$ & $\delta 1 \uparrow$ & AbsRel $\downarrow$ & $\delta 1 \uparrow$ \\
\midrule
Marigold~\cite{marigold} & 74K  & 32.9 & 80.9 & 28.0 & 78.7 & \textbf{16.2} & 82.4 & \textbf{19.6} & 80.3  \\
Metric3Dv2~\cite{hu2024metric3dv2} & 16M & 14.5 & 80.7 & 18.6 & 77.8 & 28.1 & \textbf{84.7} & 26.3 & \textbf{80.8}   \\
Depth-Anything~\cite{depthanything} & 63.5M & \textbf{12.0} & \textbf{88.4} & \textbf{14.3} & \textbf{84.7} & 46.2 & 69.0 & 46.8 & 67.8 \\

\bottomrule
\end{tabular}
}
\label{tab: fine-grained-depth}
\end{table}

\subsection{Benchmarking Different Generative Fine-Tuning Paradigms}
Several fine-tuning paradigms have been proposed for diffusion-based depth estimation. Based on
the 
network architecture, they can be divided into two categories. The first category of methods (Marigold~\cite{marigold} and DepthFM~\cite{depthfm}) concatenate the image latent and depth latent encoded by VAE~\cite{vae} encoder as the input of the UNet latent denoiser. As such, the input channels of the latent denoiser are doubled (8 input channels) to fit the expanded input. The second category methods (DMP~\cite{lee2024dmp} and GenPercept~\cite{genpercept}) drop the depth latent, so they follow the original latent denoiser's architecture (4 input channels). Based on fine-tuning paradigms, they can be divided into four categories. \textbf{(1)} Marigold~\cite{marigold} treats the initial depth latent as standard Gaussian noise and progressively denoise it with the same scheduler as the original Stable Diffusion pipeline. \textbf{(2)} DepthFM also treats the initial depth latent as standard Gaussian noise, however, the difference is that they finetune the denoiser with Flow Matching~\cite{lipman2022flowmatching} pipeline,  with auxiliary surface normal loss. \textbf{(3)} DMP~\cite{lee2024dmp} reformulates the task as a blending process, \ie, translating the image latent to depth latent with the Stable Diffusion v-prediction~\cite{ldm} learning target. \textbf{(4)} GenPercept~\cite{genpercept} further improve the efficiency of DMP~\cite{lee2024dmp} by proposing a one-step inference pipeline. Based on the amount of fine-tuned parameters, they can be divided into two categories. The first category methods (Marigold, DepthFM, GenPercept) directly fine-tune the UNet parameters. The second category method (DMP) adds LORA~\cite{hu2021lora} layers into the UNet architecture to achieve the goal of depth estimation. 

In this section, we fairly benchmark the four fine-tuning protocols by training on the Hypersim dataset (38,387 samples), with 
$480 \times 640$ resolution, $3\times10^{-5}$ learning rate, $96$ batch sizes, and $10,000$ iterations. We choose Stable Diffusion 2.1 \cite{ldm} as the base model. As shown in \cref{tab: generative_depth_benchmark}, \textbf{(1)} Marigold \cite{marigold} outperforms other protocols in outdoor benchmarks, GenPercept~\cite{genpercept} gets the best performance in indoor benchmarks. Overall, Marigold has better generation capability than GenPercept. \textbf{(2)} Fine-tuning all UNet parameters outperforms using LORA layers. (compare line 1 and line 2 on DMP) \textbf{(3)} One-step GenPercept~\cite{genpercept} can outperform multi-step DMP~\cite{lee2024dmp}. We conjecture that the RGB blending strategy proposed in DMP makes it hard to decouple image latent and depth latent during the inference stage (see Supp. Mat. for visualization). \textbf{(4)} DepthFM~\cite{depthfm} uses flow matching as the fine-tuning protocol for efficient inference (two steps). Although its performance is not comparable with Marigold, we speculate this is due to the gap between the fine-tuning flow pipeline and pre-training v-prediction pipeline in Stable Diffusion 2.1. With the rise of flow-based generative models, \eg, Stable Diffusion 3~\cite{sd3}, it may become a suitable fine-tuning strategy for flow-based models. 

\begin{table}[tb!]
\footnotesize
\setlength\tabcolsep{1.5pt}
\renewcommand{\arraystretch}{1.0} %

\centering

\caption{Benchmarking different generative finetuing paradigms on 5 zero-shot affine-invariant depth benchmarks. We mark the best results in bold and the second best underlined. }  

\resizebox{\textwidth}{!} 
{
\begin{tabular}{c|c|c|cccccccccc}
\toprule
\multirow{2}{*}{Method}  & \multirow{2}{*}{\shortstack{Train\\Samples}} & \multirow{2}{*}{\shortstack{FT\\Strategies}} &  \multicolumn{2}{c}{ NYUv2~\cite{DATA_2012_NYUv2} } & \multicolumn{2}{c}{ KITTI~\cite{DATA_2012_KITTI} } & \multicolumn{2}{c}{ ETH3D~\cite{DATA_2017_ETH3D} } & \multicolumn{2}{c}{ ScanNet~\cite{DATA_2017_ScanNet} } & \multicolumn{2}{c}{ DIODE~\cite{DATA_2019_DIODE}} \\
\cline{4-13}
& & & AbsRel $\downarrow$ & $\delta 1 \uparrow$ & AbsRel $\downarrow$ & $\delta 1 \uparrow$ & AbsRel $\downarrow$ & $\delta 1 \uparrow$ & AbsRel $\downarrow$ & $\delta 1 \uparrow$ & AbsRel $\downarrow$ & $\delta 1 \uparrow$ \\
\midrule
DMP~\cite{lee2024dmp} & 38K & LORA & 13.2 & 85.1 & 19.2 & 74.3 & 16.2 & 83.7 & 14.1  & 84.6 & 45.6 & 62.1  \\

DMP~\cite{lee2024dmp} & 38K & UNet & 10.1 & 90.6 & \underline{15.4} & \underline{80.0} & 10.0 & 91.0 & 10.9 & 89.0 & 38.2 & 68.7 \\
Marigold~\cite{marigold} & 38K & UNet & \underline{6.9} & \underline{95.0} & \textbf{13.8} & \textbf{80.7} & \textbf{7.5} & \textbf{93.7} & \underline{7.1} & \underline{94.3} & \textbf{28.7} & \textbf{74.6} \\
GenPercept~\cite{marigold} & 38K & UNet & \textbf{6.5} & \textbf{96.1} & 18.4 & 69.2 & \underline{8.5} & \underline{92.8} & \textbf{6.6} & \textbf{96.4} & \underline{32.4} & \underline{74.5} \\
DepthFM~\cite{marigold} & 38K & UNet & 10.9 & 89.5 & 19.2 & 68.8 & 12.9 & 86.3 & 11.4 & 87.7 & 33.6 & 72.4
\\

\bottomrule
\end{tabular}
}
\label{tab: generative_depth_benchmark}
\end{table}

\subsection{Benchmarking the Inference Efficiency of Depth Estimation Foundation Models}
\begin{table*}[t!]
\footnotesize
\centering
\caption{Inference latency and speed benchmark for different components and methods. `Infer Steps' indicates the minimum repeat times of the U-Net for achieving optimal results. All models are inference with $512\times512$ resolution, except CLIP~\cite{clip} ($224\times224$).}
\label{tab:inference_efficiency_bench}
\resizebox{\columnwidth}{!}{
\begin{tabular}{l|l|ccccc}
\toprule
Method & \multicolumn{1}{c|}{Components}  & Params/M & Macs/GFLOPs & Latency/s & Memory/G & Inference Steps \\ \midrule
Depth-Anything~\cite{depthanything}  & ViT-L~\cite{vit} + Head   &  335.3   &  586.0     &  0.19       & 2.24       & 1 \\ 
 Metric3Dv2~\cite{hu2024metric3dv2}  & ViT-L~\cite{vit} + Head   &  411.9    &   1014.0    &    0.60  &   2.67    & 1 \\
DSINE~\cite{bae2024dsine}  & EfficientNet B5~\cite{tan2019efficientnet} + Head    & 72.6    &  38.7  &  0.06  & 0.73   & 
 1 \\ 
\midrule
   -    & VAE-Tiny~\cite{taesd}     &   2.4   &  131.9  &   0.03     &   0.61   & 1 \\ 
   -    &  VAE~\cite{vae}    &   83.7    &   1781.2     &    0.11   &   0.65   & 1 \\ 
Geowizard~\cite{geowizard} & CLIP~\cite{clip}   &   304.0     &  77.8    &    0.04     & 1.25       & 1 \\ 
Marigold-LCM~\cite{marigold} & VAE~\cite{vae} + UNet~\cite{SEG_2015_UNET}  &  949.6      &  3138.4     &   0.29      &  5.27     & 4 \\ 
 Geowizard~\cite{geowizard}  & VAE~\cite{vae} + UNet~\cite{SEG_2015_UNET} + CLIP~\cite{clip}   &  861.2      &    9846.1   &   0.85      &  5.24     & 10 \\ 
 DepthFM~\cite{depthfm}      & VAE~\cite{vae} + UNet~\cite{SEG_2015_UNET}   & 949.6     &  2459.8     &  0.21      &  5.40   & 2 \\ 
GenPercept~\cite{genpercept} & VAE~\cite{vae} + UNet~\cite{SEG_2015_UNET}   & 949.6     &  2120.4     &   0.18  &  5.40   & 1 \\ 

GenPercept~\cite{genpercept} & VAE-Tiny~\cite{taesd} + UNet~\cite{SEG_2015_UNET}   & 868.3     &  471.1     &   0.18  &  5.40   & 1 
\\ 

\bottomrule
\end{tabular}
}
\end{table*}

Compared to discriminative models, the inference efficiency may become a bottleneck of the generative-based methods. In this section, we give detailed inference efficiency evaluation in~\cref{tab:inference_efficiency_bench}. We can see that discriminative methods have fewer parameters than generative models. The main inference consumption of the generative models happens on VAE~\cite{vae} and multiple inference steps of UNet. The last line of~\cref{tab:inference_efficiency_bench} shows that GenPercept~\cite{genpercept} can achieve comparable inference latency with Depth-Anything (ViT-Large) and a tiny VAE encoder~\cite{taesd}. In~\cref{tab: lcm_effieciency_benchmark}, we found LCM~\cite{lcm} can effectively reduce the inference steps of Marigold~\cite{marigold} while maintaining the performance. Besides, a pre-trained tiny VAE~\cite{taesd} can substitute the standard VAE~\cite{ldm} with a minimal performance loss.

\begin{table}[tb!]
\footnotesize
\setlength\tabcolsep{1.5pt}
\renewcommand{\arraystretch}{1.0} %

\centering
\caption{Benchmarking the inference efficiency of Marigold. We mark the best results in bold.
}  
\resizebox{\textwidth}{!} 
{
\begin{tabular}{c|c|c|cccccccccc}
\toprule
\multirow{2}{*}{Method}  & \multirow{2}{*}{\shortstack{VAE\\Version}} & \multirow{2}{*}{\shortstack{Infer\\Steps}} &  \multicolumn{2}{c}{ NYUv2~\cite{DATA_2012_NYUv2} } & \multicolumn{2}{c}{ KITTI~\cite{DATA_2012_KITTI} } & \multicolumn{2}{c}{ ETH3D~\cite{DATA_2017_ETH3D} } & \multicolumn{2}{c}{ ScanNet~\cite{DATA_2017_ScanNet} } & \multicolumn{2}{c}{ DIODE~\cite{DATA_2019_DIODE}} \\
\cline{4-13}
& & & AbsRel $\downarrow$ & $\delta 1 \uparrow$ & AbsRel $\downarrow$ & $\delta 1 \uparrow$ & AbsRel $\downarrow$ & $\delta 1 \uparrow$ & AbsRel $\downarrow$ & $\delta 1 \uparrow$ & AbsRel $\downarrow$ & $\delta 1 \uparrow$ \\
\midrule
Marigold~\cite{marigold} & base & 50 & \textbf{5.5} & \textbf{96.4} & \textbf{9.9} & \textbf{91.6} & {6.5} & \textbf{96.0} & \textbf{6.4} & \textbf{95.1} & \textbf{30.8} & \textbf{77.3}  \\

Marigold-LCM~\cite{marigold} & base & 4 & 6.1 & 95.8 & 10.1 & 90.6 & \textbf{6.3} & \textbf{96.0} & 6.9 & 94.7 & 30.9 & \textbf{77.3}  \\

Marigold-LCM~\cite{marigold} & tiny~\cite{taesd} & 4 & 6.9 & 95.0 & 13.8 &  80.7 & 7.5 &  93.7 &7.1 & 94.3 & 32.8 & 73.8 \\
Marigold-LCM~\cite{marigold} & tiny~\cite{taesd} & 1 & 6.6 & 95.4 & 13.0 & 83.6 & 7.8 & 93.2 & 7.0 & 94.5 & 33.3 & 73.1\\

\bottomrule
\end{tabular}
}
\label{tab: lcm_effieciency_benchmark}
\end{table}

\subsection{Benchmarking Discriminative and Generative Models on the Same Training Data}

\begin{table}[tb!]
\footnotesize
\setlength\tabcolsep{1.5pt}
\renewcommand{\arraystretch}{1.0} %

\centering
\caption{Benchmarking deterministic  and generative depth model with the same training data ($77$K).}  

\resizebox{\textwidth}{!} 
{
\begin{tabular}{c|c|c|cccccccccc}
\toprule
\multirow{2}{*}{Network}  & \multirow{2}{*}{\shortstack{Pretrain\\Style}} & \multirow{2}{*}{Backbone} &  \multicolumn{2}{c}{ NYUv2~\cite{DATA_2012_NYUv2} } & \multicolumn{2}{c}{ KITTI~\cite{DATA_2012_KITTI} } & \multicolumn{2}{c}{ ETH3D~\cite{DATA_2017_ETH3D} } & \multicolumn{2}{c}{ ScanNet~\cite{DATA_2017_ScanNet} } & \multicolumn{2}{c}{ DIODE~\cite{DATA_2019_DIODE}} \\
\cline{4-13}
& & & AbsRel $\downarrow$ & $\delta 1 \uparrow$ & AbsRel $\downarrow$ & $\delta 1 \uparrow$ & AbsRel $\downarrow$ & $\delta 1 \uparrow$ & AbsRel $\downarrow$ & $\delta 1 \uparrow$ & AbsRel $\downarrow$ & $\delta 1 \uparrow$ \\
\midrule
ViT+DPT Head & Random init & ViT-L & 21.1 & 62.5 & 27.2 & 53.1 & 23.4 & 61.1 & 19.2 & 67.4 & 32.4 & {57.7} \\
ViT+DPT Head & DINOv2~\cite{dinov2} & ViT-L & \textbf{4.9} & \textbf{97.5} & \textbf{8.5} & \textbf{94.1} & \textbf{8.1} & \textbf{97.0} & \textbf{5.1} & \textbf{97.6} &  \textbf{24.5} & \textbf{74.6}   \\
Marigold~\cite{marigold} & SD21~\cite{ldm} & UNet & 6.9 & 95.8 & 12.2 & 85.7 & 9.2 & 95.5 & 7.1 & 95.4 & 25.2 & 73.0 \\
Marigold~\cite{marigold} & SDXL~\cite{podell2023sdxl} & UNet &  6.8 & 95.8 &11.1 & 89.2 & 8.9 & 96.7  &6.3 & 96.2& \textbf{24.5} & 73.6\\
\bottomrule
\end{tabular}
}
\label{tab: depthanything_marigold_benchmark}
\end{table}

\textit{Can discriminative depth estimation models achieve competitive results with small-scale high-quality training datasets like generative-based methods?} To answer this question, we benchmark discriminative and generative geometry model with the same amount of training data and the same training strategy. Specifically, we use three training datasets, \ie, Hypersim (38,387)~\cite{2021_dataset_hypersim}, Virtual Kitti (16,790)~\cite{DATA_2016_VKITTI} and Tartanair (31,008)~\cite{2020_dataset_tartanair}, with total 77,897 samples. Both models are trained with 20,000 iterations, with a total batch size of 96 on 4 GPUs. For the discriminative depth model, we follow the network architecture of Depth-Anything~\cite{depthanything} (ViT-Large backbone pre-trained with DINOv2 and DPT~\cite{MDE_2021_DPT} head), supervised with the affine-invariant loss~\cite{depthanything}. For the generative geometry model, we choose Marigold~\cite{marigold} as our baseline.

We can see from~\cref{tab: depthanything_marigold_benchmark} that \textbf{(1)} the discriminative model is largely inferior to generative-based Marigold on all evaluation datasets without DINOv2 pre-train (line 1 \vs\  line 3). However, the discriminative model beats Marigold by a large margin when initialized with DINOv2 pre-train weight (line 2 \vs line 3); \textbf{(2)} scale-up Marigold from SD21 to SDXL brings consistent improvement in all benchmarks. We can see from~\cref{tab: proposed_depth_benchmarks} that our discriminative model trained on 77K data outperforms Metric3Dv2~\cite{hu2024metric3dv2} in all three datasets, and,  is comparable with Depth-Anything~\cite{depthanything} 
on 
two datasets (InspaceType and Infinigen). This phenomenon suggests that \textbf{high-quality fine-tuning data}, rather than large-scale training data, is indispensable for discriminative models to achieve strong generalizable performance. \textit{The qualitative visualization of discriminative models is not as good as generative models, more visualizations are available in the Supp. Mat.}

\begin{table}[tb!]
\footnotesize
\setlength\tabcolsep{1.5pt}
\renewcommand{\arraystretch}{1.0} %

\centering
\caption{Benchmarking depth estimation foundation models on more diverse benchmarks. We mark the best results in bold.}  

\resizebox{\textwidth}{!} 
{
\begin{tabular}{c|c|c|c|cccccc}
\toprule
\multirow{2}{*}{Network}  & \multirow{2}{*}{\shortstack{Pretrain\\Style}} & \multirow{2}{*}{Backbone} & \multirow{2}{*}{\shortstack{Train\\Samples}} &  \multicolumn{2}{c}{ InspaceType~\cite{wu2023inspacetype} } & \multicolumn{2}{c}{ MatrixCity~\cite{li2023matrixcity} } & \multicolumn{2}{c}{ Infinigen~\cite{infinigen} } \\
\cline{5-10}
& & & & AbsRel $\downarrow$  &$\delta 1 \uparrow$ & AbsRel $\downarrow$  & $\delta 1 \uparrow$ & AbsRel $\downarrow$  & $\delta 1 \uparrow$ \\
\midrule
Metric3Dv2~\cite{hu2024metric3dv2} & DINOv2~\cite{dinov2} & ViT-L & 16M  & 10.1 & 89.7 & \textbf{9.5} & {89.3} & 14.5 & 80.7 \\
Depth-Anything~\cite{depthanything} & DINOv2~\cite{dinov2} & ViT-L & 63.5M & \textbf{8.2} & 92.9 & {16.4} & \textbf{89.7} & 12.0 & 88.4  \\
\midrule
ViT+DPT Head & DINOv2~\cite{dinov2} & ViT-L & 77K & 8.4  & \textbf{94.0} & 28.0  & 82.4 &  \textbf{11.4} & \textbf{89.5}  \\
Marigold~\cite{marigold} & SD21~\cite{ldm} & UNet & 77K & 9.2 &  92.7&17.0  & 82.9 &  14.1 & 83.9    \\
\bottomrule
\end{tabular}
}
\label{tab: proposed_depth_benchmarks}
\end{table}

\section{Benchmarking Surface Normal Estimation Foundation Models}
\subsection{A Brief Overview of SOTA Methods}
DSINE~\cite{bae2024dsine} and Metric3Dv2~\cite{hu2024metric3dv2} are two representative discriminative surface estimation models, which leverage the geometry priors from two distinct perspectives. DSINE leverages two forms of inductive bias: \textbf{(1)} per-pixel ray direction, and \textbf{(2)} the relationship between the neighboring surface normal, to learn a generalizable surface normal estimator. Metric3Dv2~\cite{hu2024metric3dv2} proposes to optimize the surface normal map by distilling diverse data knowledge from the estimated metric depth. Different from discriminative models, GeoWizard~\cite{geowizard} is a generative surface normal estimator without using any inductive bias from the geometry priors. It purely relies on pre-trained diffusion priors to estimate the surface normal map. \cref{table:official_normal_benchmark} summarizes their performance on six benchmarks. The Mushroom~\cite{DATA_2024_MUSHROOM}  (indoor), T\&T~\cite{DATA_2017_Tanks_and_Temples} (outdoor), and Infinigen~\cite{infinigen} (wild) datasets are constructed by us to add more diverse scenes with accurate surface normal labels in the evaluation benchmarks. We can see that Metric3Dv2~\cite{hu2024metric3dv2} outperform DSINE~\cite{bae2024dsine} and GeoWizard~\cite{geowizard} in most datasets.  Note it is an unfair comparison since \textbf{(1)} Metric3Dv2~\cite{hu2024metric3dv2} is trained on 16M images, while DSINE is trained on 160K images, and GeoWizard is trained on 280K images. \textbf{(2)} DSINE use a much smaller backbone, EfficientNet-B5~\cite{tan2019efficientnet}, while Metric3Dv2~\cite{hu2024metric3dv2} employs the ViT-Large~\cite{vit} backbone, pretrained using DINOv2 with registers~\cite{darcet2023vitneedreg}.

\subsection{Benchmarking Discriminative and Generative Models on the Same Training Data}
In this section, we fairly benchmark discriminative DSINE~\cite{bae2024dsine} and several representative generative geometry models, \ie, Marigold~\cite{marigold}, DMP~\cite{lee2024dmp}, GenPercept~\cite{genpercept}, and DepthFM~\cite{depthfm}, with 5 training datasets, Hypersim~\cite{2021_dataset_hypersim} ($38,387$), Tartanair~\cite{2020_dataset_tartanair} ($31,008$), Virtual Kitti~\cite{DATA_2016_VKITTI} ($16,790$), BlendedMVS~\cite{2020_dataset_blendedmvs} ($17,819$), ClearGrasp~\cite{2020_dataset_cleargrasp} ($22,720$), a total of $126,724$ samples. For generatative-based models, we represent the output surface normals as unit vectors. We follow DSINE~\cite{bae2024dsine} to represent the outputs of discriminative-based model as axis-angles with three degrees of freedom. All models are trained with $20,000$ iterations, $96$ batch sizes, $480\times 640$ resolution on 4 A800 GPUs. All generative-based models use $3\times10^{-5}$ learning rate. For discriminative model, we follow DSINE~\cite{bae2024dsine} to use $3\times10^{-5}$ learning rate for the backbone and $3\times10^{-4}$ learning rate for the decoder.

We can see from~\cref{table: normal_benchmark} that \textbf{(1)} DSINE can scale up the performance by using ViT-Large backbone with DINOv2 pretrain (compared with ImageNet pretrained Efficient-B5 backbone). \textbf{(2)} For generative-based fine-tuning protocols, DepthFM~\cite{depthfm} outperforms other paradigms in most benchmarks. We attribute this to the decoder supervision during the training. Paradigms that requires multi-step denoising inference steps, \eg, Marigold~\cite{marigold} and DMP~\cite{lee2024dmp}are not suitable to perform decoder supervision during the training. \textbf{(3)} Discriminative models, equipped with inductive bias, outperform generative-based models with only diffusion priors. It is promising to inject inductive bias into the diffusion-based models, as such, the surface normal estimator can effectively leverage the diffusion priors and inductive bias to boost the performance. \textbf{(4)} DSINE (ViT-Large in~\cref{table: normal_benchmark}) trained with $120$K samples achieves comparable performance with Metric3Dv2 trained with $16$M samples (\cref{table:official_normal_benchmark}). The results verify the point that data-quality is more important than the data-scale.

\begin{table*}[t!]
\footnotesize
\setlength\tabcolsep{1.5pt}
\renewcommand{\arraystretch}{1.0} %
\caption{Quantitative evaluation of the generalization capabilities possessed by \textbf{
various 
methods with official released weights.} For each metric, the best results are bolded. Discriminative methods are colored in \colorbox{blue!20}{blue} while generative ones in \colorbox{green!20}{green}.
}

\begin{center}
\resizebox{\columnwidth}{!}{%
\begin{tabular}{l|cc|ccccc|cc|ccccc|cc|ccccc}
\toprule
\multirow{2}{*}{Method} 
& \multicolumn{7}{c|}{NYU v2~\cite{DATA_2012_NYUv2}}
& \multicolumn{7}{c|}{ScanNet~\cite{DATA_2017_ScanNet}}
& \multicolumn{7}{c}{Sintel~\cite{DATA_2012_SINTEL}} \\
\cline{2-22}
& mean & med & {\scriptsize $5.0^{\circ}$} & {\scriptsize $7.5^{\circ}$} & {\scriptsize $11.25^{\circ}$} & {\scriptsize $22.5^{\circ}$} & {\scriptsize $30^{\circ}$} 
& mean & med & {\scriptsize $5.0^{\circ}$} & {\scriptsize $7.5^{\circ}$} & {\scriptsize $11.25^{\circ}$} & {\scriptsize $22.5^{\circ}$} & {\scriptsize $30^{\circ}$} 
& mean & med & {\scriptsize $5.0^{\circ}$} & {\scriptsize $7.5^{\circ}$} & {\scriptsize $11.25^{\circ}$} & {\scriptsize $22.5^{\circ}$} & {\scriptsize $30^{\circ}$} \\
\midrule
\rowcolor{blue!10}Metric3Dv2~\cite{hu2024metric3dv2} & 
\textbf{13.5} & \textbf{6.7} & \textbf{40.1} &  \textbf{53.5} &  \textbf{65.9} & \textbf{82.6}  & \textbf{87.7} &
\textbf{11.8} & \textbf{5.5} & \textbf{46.6} & \textbf{60.7} &  \textbf{71.6}  &  \textbf{85.4} &  \textbf{89.7} & 
\textbf{22.8} & \textbf{14.2} & \textbf{18.4} & \textbf{28.5} & \textbf{41.6} & \textbf{66.7} & \textbf{75.8} \\
\rowcolor{blue!10}DINSE~\cite{bae2024dsine} & 
16.4 & 8.4 & 32.8 & 46.3 & 59.6 & 77.7 & 83.5 &
18.3 & 9.3 & 27.1 & 42.0 & 56.3 & 75.0 & 81.2 &
32.0 & 23.9 & 9.0 & 15.0 & 23.8 & 47.5 & 59.4 \\ 
\midrule
\rowcolor{green!10}Geowizard~\cite{geowizard} & 
19.8 & 11.2 & 18.0 & 32.7 & 50.2 &73.0 & 79.9 & 
21.1 & 11.9 & 15.9 & 29.7 & 47.4& 70.7& 77.8& 
36.1 & 28.4 & 4.1 & 8.6 & 16.9 & 39.8 & 52.5 \\
\midrule
\multirow{2}{*}{Method}
& \multicolumn{7}{c|}{MuSHRoom Subset~\cite{DATA_2024_MUSHROOM} (Indoor) }
& \multicolumn{7}{c|}{T\&T Subset~\cite{DATA_2017_Tanks_and_Temples} (Outdoor)}
& \multicolumn{7}{c}{Infinigen Subset~\cite{infinigen} (Wild)} \\
\cline{2-22}
& mean & med & {\scriptsize $5.0^{\circ}$} & {\scriptsize $7.5^{\circ}$} & {\scriptsize $11.25^{\circ}$} & {\scriptsize $22.5^{\circ}$} & {\scriptsize $30^{\circ}$} 
& mean & med & {\scriptsize $5.0^{\circ}$} & {\scriptsize $7.5^{\circ}$} & {\scriptsize $11.25^{\circ}$} & {\scriptsize $22.5^{\circ}$} & {\scriptsize $30^{\circ}$} 
& mean & med & {\scriptsize $5.0^{\circ}$} & {\scriptsize $7.5^{\circ}$} & {\scriptsize $11.25^{\circ}$} & {\scriptsize $22.5^{\circ}$} & {\scriptsize $30^{\circ}$} \\
\midrule
\rowcolor{blue!10}Metric3Dv2~\cite{hu2024metric3dv2} & 
\textbf{14.3} & \textbf{7.9} & \textbf{31.9} & \textbf{48.1}  & \textbf{61.8}  & \textbf{81.7}  & \textbf{87.2} &
22.3 & 14.1 & 19.2 & 31.4  & 43.0  & 64.8  & 73.5 &
\textbf{32.6} & \textbf{27.3} & \textbf{5.1} & \textbf{10.1} & \textbf{17.8} & \textbf{41.3} & \textbf{54.4} \\
\rowcolor{blue!10}DINSE~\cite{bae2024dsine} & 
14.8 & 8.6 & 28.1 & 44.6 & 59.7 & 80.4 & 87.0 & 
\textbf{17.3} & \textbf{11.0} & \textbf{24.2} & \textbf{37.3} & \textbf{50.6} & \textbf{74.1} & \textbf{82.4} &
35.9 & 32.6 & 2.1 & 4.6 & 9.8 & 30.5 & 45.1 \\ 
\midrule
\rowcolor{green!10}Geowizard~\cite{geowizard} & 
 16.5 & 10.7 & 14.7 &  30.5 &  52.5 & 79.6  & 86.2 &
 20.8 & 13.4 & 10.7 & 23.4  & 42.2  & 70.3  & 78.5 &
 36.2 & 32.0 & 1.8 & 4.0 & 8.86  &  30.8 & 46.2   \\
\bottomrule
\end{tabular}
}
\end{center}
\label{table: normal_benchmark}
\end{table*}

\begin{table*}[t!]
\footnotesize
\setlength\tabcolsep{1.5pt}
\caption{Quantitative evaluation of the generalization capabilities \textbf{with the same training data} on different benchmarks. The best results are bolded. `EB5' indicates ImageNet~\cite{imagenet} pre-trained EfficientNet-B5~\cite{tan2019efficientnet}. `ViT-L' indicates DINOv2~\cite{dinov2} pre-trained ViT-Large. The best results are \textbf{bolded}.  The best results of generativate-based models are \underline{underlined}.}
\begin{center}
\resizebox{\columnwidth}{!}{%
\begin{tabular}{l|c|cc|ccccc|cc|ccccc|cc|ccccc}
\toprule
\multirow{2}{*}{Method} & \multirow{2}{*}{Backbone} 
& \multicolumn{7}{c|}{NYUv2~\cite{DATA_2012_NYUv2}}
& \multicolumn{7}{c|}{ScanNet~\cite{DATA_2017_ScanNet}}
& \multicolumn{7}{c}{Sintel~\cite{DATA_2012_SINTEL}} \\
\cline{3-23}
& &mean & med & {\scriptsize $5.0^{\circ}$} & {\scriptsize $7.5^{\circ}$} & {\scriptsize $11.25^{\circ}$} & {\scriptsize $22.5^{\circ}$} & {\scriptsize $30^{\circ}$} 
& mean & med & {\scriptsize $5.0^{\circ}$} & {\scriptsize $7.5^{\circ}$} & {\scriptsize $11.25^{\circ}$} & {\scriptsize $22.5^{\circ}$} & {\scriptsize $30^{\circ}$} 
& mean & med & {\scriptsize $5.0^{\circ}$} & {\scriptsize $7.5^{\circ}$} & {\scriptsize $11.25^{\circ}$} & {\scriptsize $22.5^{\circ}$} & {\scriptsize $30^{\circ}$} \\
\midrule
\rowcolor{blue!10}DSINE~\cite{bae2024dsine} & EB5~\cite{tan2019efficientnet} &
19.2 & 10.0 & 27.1 & 40.1 & 53.9 & 73.8 & 80.1 & 
17.3 & 11.0 & 24.2 & 37.3 & 50.6 & 74.1 & 82.4 &
35.9 & 32.6 & 2.1 & 4.6 & 9.8 & 30.5 & 45.1 \\ 
\rowcolor{blue!10}DSINE~\cite{bae2024dsine} & ViT-L~\cite{dinov2} &
\textbf{16.2} & \textbf{8.2} & \textbf{32.8} & \textbf{46.9} & \textbf{60.6} & \textbf{78.5} & \textbf{84.1} & 
\textbf{16.1} & \textbf{7.4} & \textbf{34.5} & \textbf{50.6} & \textbf{63.8} & \textbf{79.4} & \textbf{84.3} &
\textbf{24.6} & \textbf{16.1} & \textbf{11.1} & \textbf{21.1} & \textbf{35.1} & \textbf{63.8} & \textbf{74.1} \\
\rowcolor{green!10}GenPercept~\cite{genpercept} & UNet~\cite{SEG_2015_UNET} & 
\underline{17.8} & 9.5  &  24.1 &  40.5 & 55.9 & 75.7  & 82.2 &
\underline{18.5} & 9.4  &  23.0 & 40.2  & 56.7 & 75.4  & \underline{81.3} &
38.6 & 27.1 &  4.5  &  9.1  & 18.0 & 42.5  & 54.2 \\
\rowcolor{green!10}Marigold~\cite{marigold} & UNet~\cite{SEG_2015_UNET} & 
20.2 & 10.9 & 21.8 &  36.0 &   51.2 & 72.8 & 79.4 &
20.5 & 10.3 & 19.7 &  36.0 &   53.5 & 73.6  & 79.2 &
41.3 & 28.7 & 5.5 & 11.1 & 19.7 & 40.9 & 51.7 \\
\rowcolor{green!10}DMP~\cite{lee2024dmp} & UNet~\cite{SEG_2015_UNET} &
21.9 & 11.3 & 19.7 & 34.2 & 49.7 &71.1 & 77.6 &
22.5 & 11.2 & 17.6 & 32.5 & 50.3 & 71.2 & 76.9 & 
45.0 & 39.3 & 4.2 & 7.9 & 13.8 & 29.9 & 39.0 \\
\rowcolor{green!10}DepthFM~\cite{depthfm} & UNet~\cite{SEG_2015_UNET} &
\underline{17.8} & \underline{9.3} & \underline{27.7} & \underline{41.9} & \underline{56.5} & \underline{76.7} & \underline{82.5} &
\underline{18.5} & \underline{8.6} & \underline{28.4} & \underline{44.7} & \underline{58.8} & \underline{75.6} & 81.0 &
\underline{34.1} & \underline{25.8} & \underline{7.1} & \underline{13.5} & \underline{22.0} & \underline{44.8} & \underline{55.7} 
\\
\midrule
\multirow{2}{*}{Method} & \multirow{2}{*}{Backbone} 
& \multicolumn{7}{c|}{MuSHRoom Subset~\cite{DATA_2024_MUSHROOM} (Indoor) }
& \multicolumn{7}{c|}{T\&T Subset~\cite{DATA_2017_Tanks_and_Temples} (Outdoor)}
& \multicolumn{7}{c}{Infinigen Subset~\cite{infinigen} (Wild)} \\
\cline{3-23}
& & mean & med & {\scriptsize $5.0^{\circ}$} & {\scriptsize $7.5^{\circ}$} & {\scriptsize $11.25^{\circ}$} & {\scriptsize $22.5^{\circ}$} & {\scriptsize $30^{\circ}$} 
& mean & med & {\scriptsize $5.0^{\circ}$} & {\scriptsize $7.5^{\circ}$} & {\scriptsize $11.25^{\circ}$} & {\scriptsize $22.5^{\circ}$} & {\scriptsize $30^{\circ}$} 
& mean & med & {\scriptsize $5.0^{\circ}$} & {\scriptsize $7.5^{\circ}$} & {\scriptsize $11.25^{\circ}$} & {\scriptsize $22.5^{\circ}$} & {\scriptsize $30^{\circ}$} \\
\midrule
\rowcolor{blue!10}DSINE~\cite{bae2024dsine} & EB5~\cite{tan2019efficientnet} &
17.9 & 10.1 & 23.9 & 38.9 & 53.8 & 75.8 & 82.5 & 
21.7 & 15.4 & 13.4 & 25.7 & 39.0 & 64.4 & 75.2 &
36.5 & 32.7 & 2.0 & 4.4 & 9.3 & 29.8 & 44.8 \\ 
\rowcolor{blue!10}DSINE~\cite{bae2024dsine} & ViT-L~\cite{dinov2} &
 \textbf{12.8} & \textbf{6.9} & \textbf{34.6} & \textbf{53.5} & \textbf{67.9} & \textbf{84.9} & \textbf{89.6} & 
 \textbf{18.8} & \textbf{11.2} & \textbf{22.5}  & \textbf{36.3} & \textbf{49.9} & \textbf{71.1} & \textbf{79.5} & 
\textbf{33.6} & \textbf{28.7} & \textbf{2.7} & \textbf{5.9} & \textbf{13.1} & \textbf{38.3} & \textbf{52.2} \\
\rowcolor{green!10}GenPercept~\cite{genpercept} & UNet~\cite{SEG_2015_UNET} & 
 \underline{15.0} & \underline{7.9} & \underline{32.4} & \underline{48.0} & \underline{62.7} & \underline{80.6} & \underline{86.3} & 
 27.4 & 14.0  &17.8 & 30.1 & 43.1 & 63.7 & 71.0 & 
 38.8 & 33.5 & 2.1 &  4.9 & 10.5 & 31.2 & 44.3 \\
\rowcolor{green!10}Marigold~\cite{marigold} & UNet~\cite{SEG_2015_UNET} & 
 17.7 & 9.9 & 19.6 & 36.8 & 55.7 & 77.0 & 83.1 & 
 29.1 & 14.6 & 14.4& 26.0 & 40.3 & 63.0 & 70.2 & 
 39.2 & 34.0 & 2.4 & 5.4 & 11.5 & 31.4 & 43.9\\
\rowcolor{green!10}DMP~\cite{lee2024dmp} & UNet~\cite{SEG_2015_UNET} &
 20.4 & 10.0 & 19.3 & 36.0 & 55.2 & 73.8 & 79.1  & 
 27.7 & 17.9 & 9.1 & 17.2  & 31.9 &57.9  & 66.1  & 
 43.1 & 38.1 & 1.7 & 4.0 & 9.5 & 26.3 & 38.1 \\
\rowcolor{green!10}DepthFM~\cite{depthfm} & UNet~\cite{SEG_2015_UNET} &
 17.0 & 9.0 & 25.9 & 42.5 & 58.9 & 77.4 & 82.4 & 
 \underline{22.1} & \underline{13.9} & \underline{14.2} & \underline{26.7} & \underline{42.1} & \underline{65.4} & \underline{74.1} & 
 \underline{31.9} &  \underline{27.9} & \underline{2.5} & \underline{5.4} & \underline{11.6} & \underline{38.3} & \underline{54.0}
\\
\bottomrule
\end{tabular}
}
\end{center}
\label{table:official_normal_benchmark}
\end{table*}

\section{Benchmarking Geometric Correspondence}

\textit{Can current monocular geometry estimation models improve the 3D awareness of the original representation models, \eg, DINOv2 and Stable Diffusion}? To answer the question, we follow Probe3D~\cite{probe3d} by using geometric correspondence estimation, since 3D awareness implies consistency of representations across different views. Specifically, given two views of the same scene, geometric correspondence estimation needs to identify pixels across views that depict the same point in 3D space. We extract feature maps from either trained monocular geometry models or representation models, \eg, DINOv2, and directly compute correspondence between the dense feature maps of different views. We use Paired ScanNet~\cite{DATA_2017_ScanNet} for scene evaluation and NAVI wild set~\cite{jampani2024navi} for object evaluation. Following~\cite {probe3d}, we report the correspondence recall, \ie, the percentage of correspondence that falls within some defined distance.
\begin{figure*}
    \centering
    \includegraphics[width=1.0\linewidth]{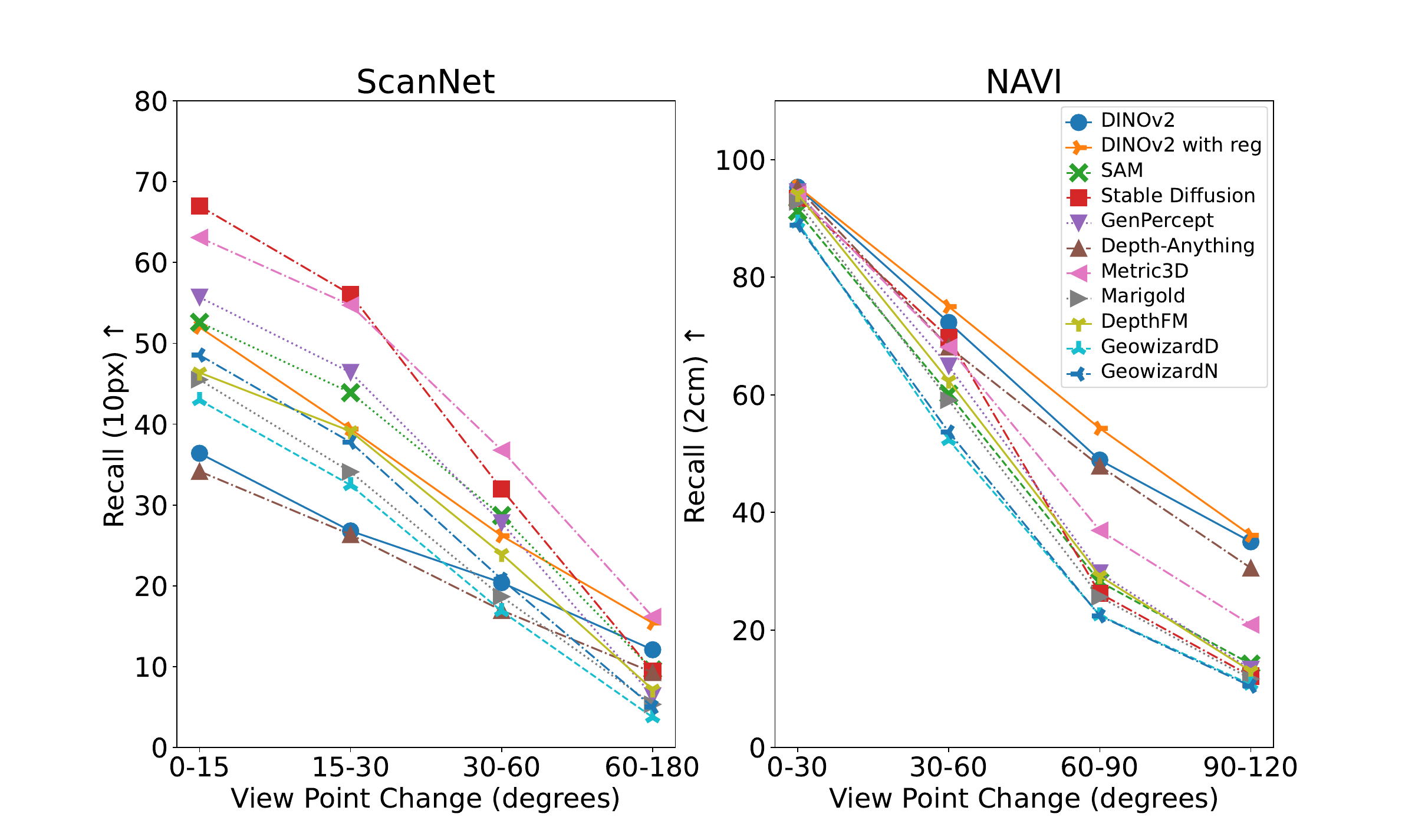}
    \caption{Geometry correspondences evaluation. `GeowizardD', `GeowizardN' indicate depth and normal features from Geowizard.}    
    \label{fig:geo_corr}
\end{figure*}
We can see from  \cref{fig:geo_corr} that: 
\begin{enumerate}
    \item 


Depth-Anything, pretrained with DINOv2~\cite{dinov2} and fine-tuned on $77$K training samples, is comparable to the original DINOv2; Metric3Dv2 pretrained with DINOv2-reg~\cite{darcet2023vitneedreg} outperforms original DINOv2-reg on ScanNet and inferior to original DINOv2-reg on NAVI dataset. While generative-based models, \ie, Marigold, DepthFM, GenPercept, and Geowizard, are inferior to the original Stable Diffusion~\cite{ldm} model on both datasets.


\item 
All models struggle with larger view changes, while generative-based models see a larger drop. In general, monocular geometry estimation models are not 3D-consistent with large viewpoints and thus not yet good enough to encode the 3D structure of the real-world scenario.

\end{enumerate}

\section{Conclusion and Discussion}
In this work, we present the \textit{first} large-scale benchmarking of discriminative and generative geometry estimation foundation models with diverse evaluation datasets. We identify that with a strong pre-train model, either Stable Diffusion or DINOv2, the fine-tuning data quality is a more important factor than fine-tuning data scale and model architecture to achieve generalizable geometry estimation. We believe that this benchmarking study can provide strong baselines for unbiased comparisons in geometry estimation studies. Limitations and future works are discussed in the Supp. Mat.

\clearpage
\appendix
\def\x{{\ensuremath \times}}

\section{Appendix}

\textbf{Benchmarking}: We introduce new benchmarks with diverse scenes and high-quality annotations to overcome limitations in scene diversity and label quality of existing benchmarks.

\textbf{Findings}: The study reveals that discriminative models pre-trained on large data can outperform generative models with less but high-quality synthetic data, suggesting the importance of data quality over quantity and architecture.

\subsection{Depth Estimation Visualization of Different Methods}
\begin{figure}[h]
    \centering
    \includegraphics[width=0.999\linewidth]{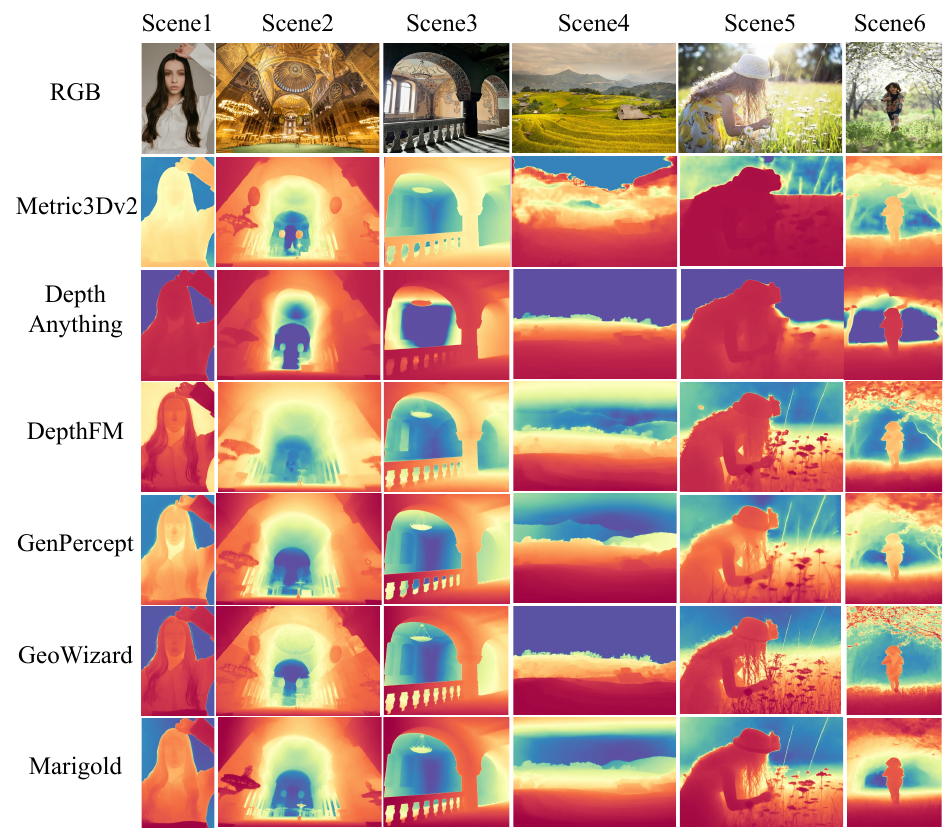}
    \caption{Visualization of different depth estimation methods.}
    \label{fig:depth_vis_1}
\end{figure}

\begin{figure}[h]
    \centering
    \includegraphics[width=0.999\linewidth]{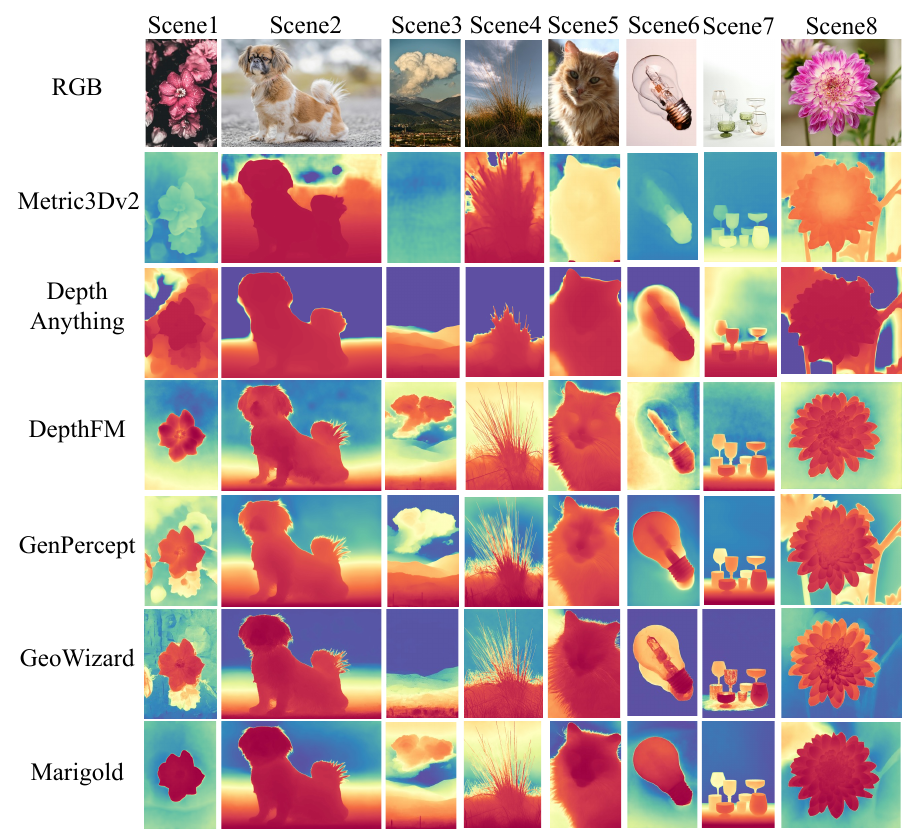}
    \caption{Visualization of different depth estimation methods.}
    \label{fig:depth_vis_2}
\end{figure}

We visualize the depth estimation results in~\cref{fig:depth_vis_1} and ~\cref{fig:depth_vis_2}.

\subsection{Surface Normal Estimation Visualization of Different Methods}
\begin{figure}
    \centering
    \includegraphics[width=0.999\linewidth]{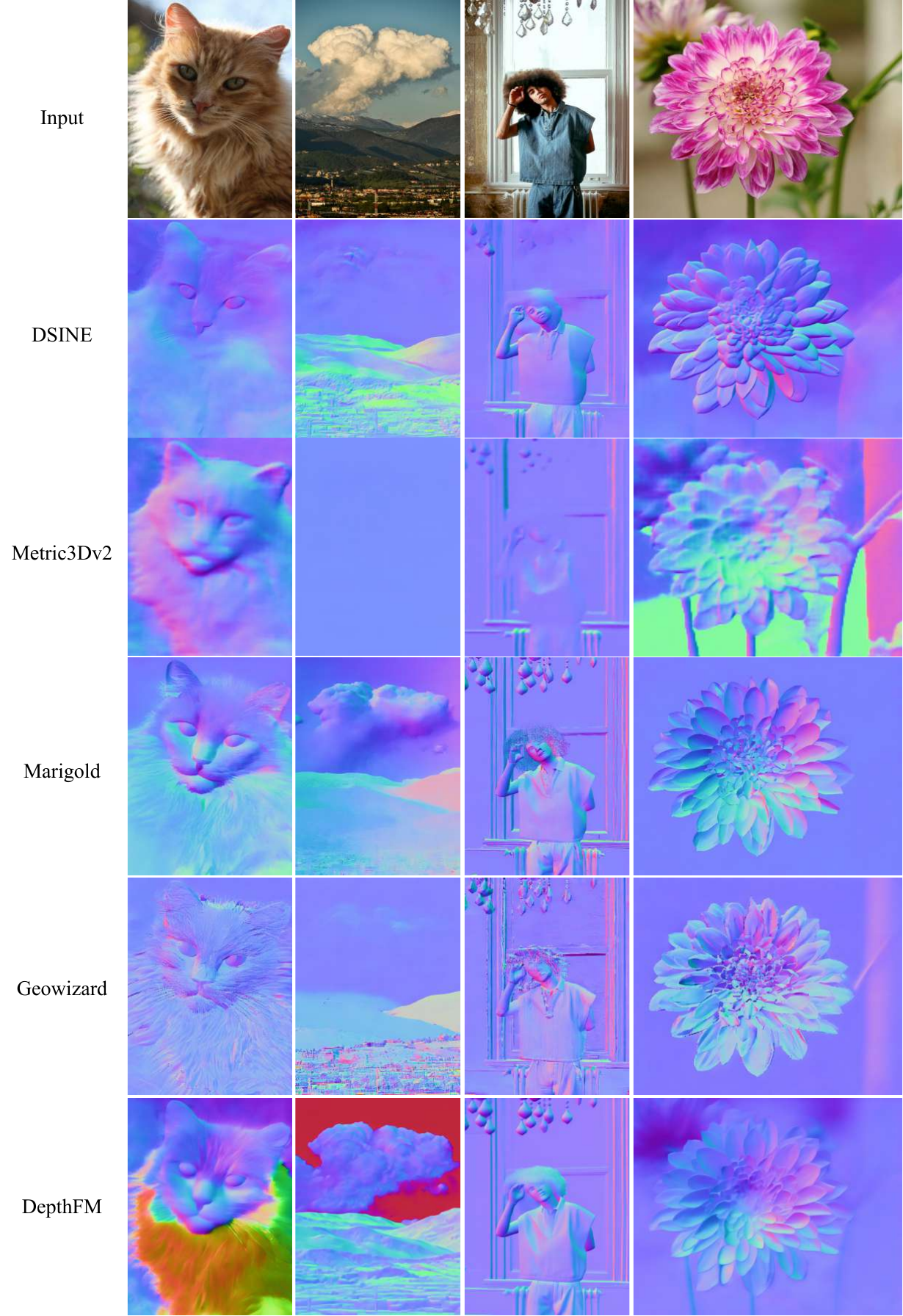}
    \caption{Visualization of different surface normal estimation methods.}
    \label{fig:normal_vis_1}
\end{figure}

\begin{figure}
    \centering
    \includegraphics[width=0.999\linewidth]{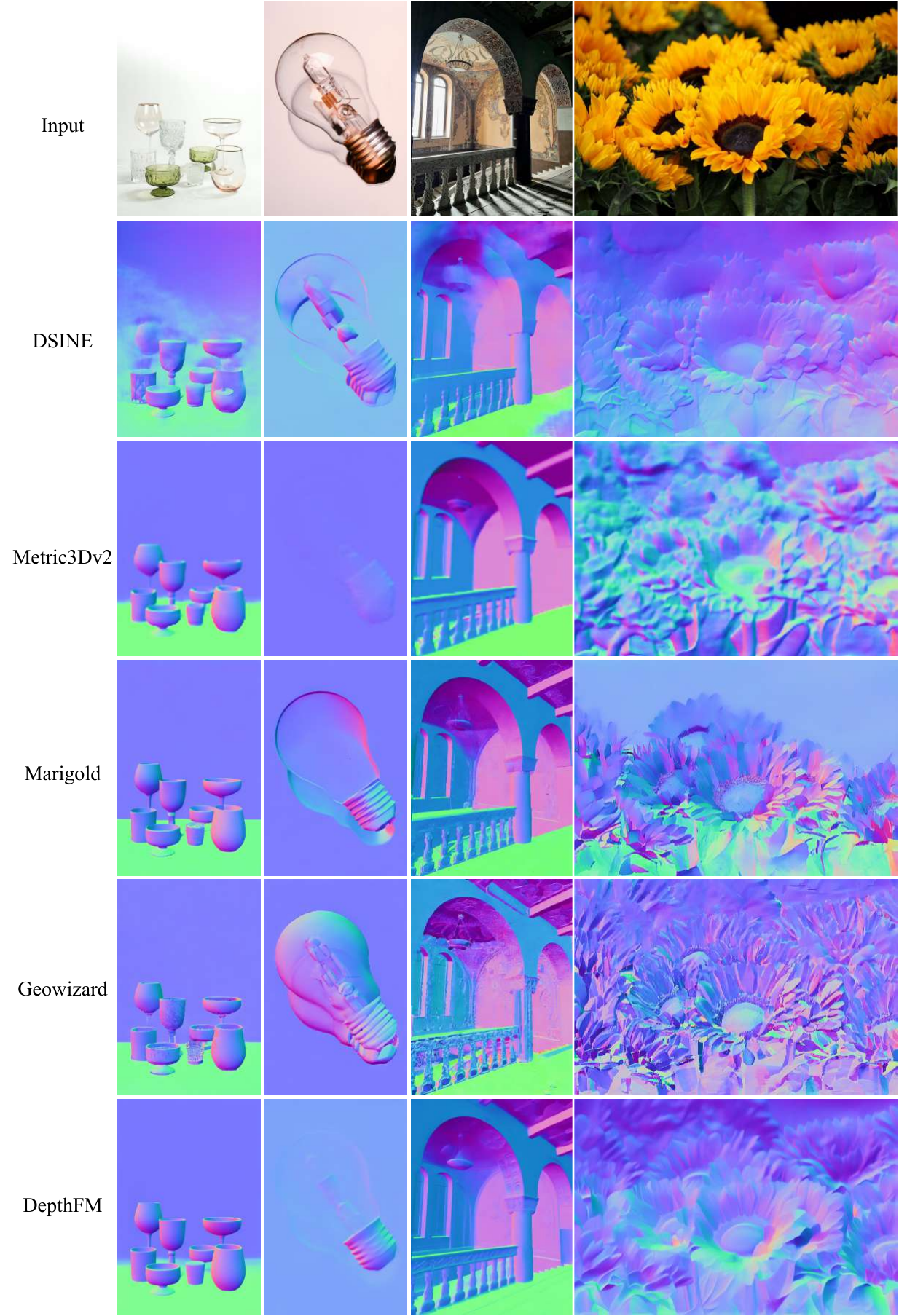}
    \caption{Visualization of different surface normal estimation methods.}
    \label{fig:normal_vis_2}
\end{figure}

We visualize the surface normal estimation results in~\cref{fig:normal_vis_1} and ~\cref{fig:normal_vis_2}.

\subsection{Correspondence Estimation Results}
We give more detailed correspondence estimation results in~\cref{tab:corr_estimation} for reference. Note that we find that multi-step inference, \textit{e.g.}, 10 steps, can improve the performance of Stable Diffusion in correspondence estimation tasks. Metric3Dv2~\cite{hu2024metric3dv2} employs DINOv2 with registers~\cite{darcet2023vitneedreg} as the backbone, which has higher performance than DINOv2 without registers~\cite{dinov2}.

\begin{table*}[h]
  \newcommand{\rowheader}{\rowcolor{blue!10}}
  \newcommand{\rowheadergray}{\rowcolor{gray!10}}
  \newcommand{\rowheadergreen}{\rowcolor{green!10}}
  \newcommand{\hlc}{\cellcolor{Orange!20}}
  \renewcommand{\arraystretch}{1.0} %
  \centering
  \caption{\textbf{Correspondence Estimation Results.} The results are presented for features extracted at different layers with performance binned based on the viewpoint variation for the image pair. `DA' indicates Depth-Anything. `DA (77K)' indicates Depth-Anything trained with only 77K synthetic data. `SD10' indicates Stable Diffusion model inference 10 steps. `MIX' indicates using a mixture of datasets during the training. The higher the recall in the table, the better the performance.}  
  \label{tab:corr_estimation}
  \setlength\tabcolsep{3.0pt}
  \scriptsize
\resizebox{1.\textwidth}{!}
{
  \begin{tabularx}{1.071\linewidth}{Xlll@{\hskip 10pt} rrrr@{\hskip 10pt} rrrr@{\hskip 10pt} rrrr@{\hskip 10pt} rrrr}
    \toprule
    & & &
    & \multicolumn{4}{c}{Spair-71k}
    & \multicolumn{4}{c}{Paired ScanNet}
    & \multicolumn{4}{c}{NAVI}
    \\
    \cmidrule(lr){5-8} 
    \cmidrule(lr){9-12}
    \cmidrule(lr){13-16}

    \textbf{Model} & \textbf{Architecture} & \textbf{Dataset}  & \textbf{Layers} &
    $d{=}0$ & $d{=}1$ & $d{=}2$ & all &
    $\theta_{0}^{15}$ & $\theta_{15}^{30}$ & $\theta_{30}^{60}$ & $\theta_{60}^{180}$ &
    $\theta_{0}^{30}$ & $\theta_{30}^{60}$ & $\theta_{60}^{90}$ & $\theta_{90}^{120}$  \\
    \midrule
\rowheadergray \multicolumn{16}{l}{\textit{\textbf{Pre-train Models}}} \\

DINOv2~\cite{dinov2}           & ViT-L14   & LVD     &  Block0  &  8.5  &  6.2  &  5.3  &  7.5    & 17.2 & 14.1 & 10.1 &  4.7   & 66.2 & 37.2 & 19.6 & 11.5 \\

DINOv2~\cite{dinov2}                   & ViT-L14   & LVD     &  Block1  & 25.0  & 14.0  & 10.8  & 19.3    & 29.0 & 20.8 & 13.5 &  5.2   & 92.1 & 57.9 & 25.6 & 12.8 \\

DINOv2~\cite{dinov2}                      & ViT-L14   & LVD     &  Block2  & 53.9  & 34.6  & 31.6  & 44.5    & 35.2 & 24.1 & 16.3 &  6.6   & 95.3 & 70.0 & 35.4 & 18.5 \\

DINOv2~\cite{dinov2}                        & ViT-L14   & LVD     &  Block3  & 62.8  & 53.3  & 54.2  & 57.2    & 36.5 & 27.0 & 20.8 & 12.2   & 92.2 & 72.3 & 48.9 & 35.0 \\

DINOv2~\cite{darcet2023vitneedreg}           & ViT-L14+reg   & LVD     &  Block0  &  12.2&   8.8&   8.1&  10.4   & 14.0& 14.2& 11.4&  5.0   &  79.9& 40.8 & 24.5& 13.6 \\

DINOv2~\cite{darcet2023vitneedreg}                       & ViT-L14+reg   & LVD     &  Block1  & 41.2&  22.8&  17.1&  32.0     & 52.0& 39.4& 23.7&  9.1   & 95.4& 65.6& 32.7& 15.8 \\

DINOv2~\cite{darcet2023vitneedreg}                        & ViT-L14+reg   & LVD     &  Block2  & 64.2&  45.9&  42.4&  55.0     & 50.6& 39.3& 26.2& 12.0   & 95.2& 75.0& 49.1& 28.6  \\

DINOv2~\cite{darcet2023vitneedreg}                        & ViT-L14+reg   & LVD     &  Block3  & 59.3&  53.2&  54.9&  55.0    & 45.0& 35.4& 26.1& 15.4   & 88.6& 71.2& 54.3& 36.1  \\

SAM~\cite{SAM}              & ViT-L16   & SA-1B   &  Block0  &   9.9 &   6.1 &   5.4 &   8.0   & 14.5 &  9.9 &  7.5 & 3.5   & 78.0 & 43.3 & 20.4 & 11.4  \\

SAM~\cite{SAM}                            & ViT-L16   & SA-1B   &  Block1  &  22.6 &  15.8 &  12.5 &  18.3   & 37.2 & 29.7 & 19.7 & 6.2   & 86.4 & 52.0 & 23.8 & 12.5  \\

SAM~\cite{SAM}                             & ViT-L16   & SA-1B   &  Block2  &  34.8 &  23.1 &  17.0 &  28.2   & 47.6 & 40.4 & 27.3 & 8.7   & 91.2 & 60.1 & 28.2 & 14.2  \\

SAM~\cite{SAM}                      & ViT-L16   & SA-1B   &  Block3  &  30.2 &  18.1 &  13.0 &  24.1   & 52.6 & 43.9 & 28.7 & 9.6   & 88.5 & 57.6 & 26.9 & 13.5  \\

SD10~\cite{ldm}  & UNet      & LAION  & Block0 & 13.2 &  5.3 &  3.5 &  9.2   & 10.8 &  5.4 &  3.2 & 1.3    & 75.1 & 32.5 & 16.6 &  7.4 \\
SD10~\cite{ldm}  & UNet      & LAION  & Block1 & 58.6 & 36.4 & 28.6 & 47.8   & 67.0 & 56.1 & 32.0 & 8.7    & 93.4 & 59.7 & 26.2 & 11.4 \\
SD10~\cite{ldm}  & UNet      & LAION  & Block2 & 24.0 & 16.8 & 13.4 & 20.2   & 61.4 & 49.5 & 28.4 & 9.4    & 79.0 & 42.5 & 22.3 & 12.2 \\
SD10~\cite{ldm}  & UNet      & LAION  & Block3 &  4.6 &  4.3 &  4.4 &  4.3   & 17.2 & 12.8 &  8.9 & 5.0    & 35.3 & 22.9 & 15.2 & 11.0 \\
\midrule \rowheader \multicolumn{16}{l}{\textit{\textbf{Deterministic Geometry Foundation Models}}} \\

MiDaS~\cite{MDE_2021_DPT}              & ViT-L16   & MIX 6 &  Block0    & 15.6 & 10.2 &  8.7 & 13.0   & 50.3 & 39.0 & 24.4 & 11.2   & 79.0 & 49.1 & 25.0 & 14.5  \\

MiDaS~\cite{MDE_2021_DPT}              & ViT-L16   & MIX 6 &  Block1    & 27.3 & 22.8 &  23.2 & 24.5   & 56.4 & 47.4 & 31.6 & 13.9   & 83.2 & 56.0 & 32.1 & 21.6  \\

MiDaS~\cite{MDE_2021_DPT}      & ViT-L16   & MIX 6 &  Block2    & 28.2 & 23.4 &  25.1 & 25.5   & 55.5 & 46.0 & 30.8 & 14.3   & 82.2 & 56.3 & 33.1 & 22.9  \\

MiDaS~\cite{MDE_2021_DPT}     & ViT-L16   & MIX 6 &  Block3    & 25.8 & 21.3 &  23.6 & 23.4   & 52.4 & 42.1 & 27.6 & 13.1   & 79.6 & 53.0 & 31.4 & 21.6  \\

DA~\cite{depthanything}              & ViT-L16   & MIX & Block0 &  8.0 &  6.1 &  5.3 &  6.8     & 21.4 & 17.5 & 12.2 & 5.4   & 66.1 & 35.6 & 20.6 & 12.5 \\

DA~\cite{depthanything}              & ViT-L16   & MIX & Block1 & 24.4 & 13.8 & 11.1 & 19.4     & 34.2 & 26.4 & 17.0 & 6.1   & 92.4 & 55.9 & 27.7 & 14.0 \\

DA~\cite{depthanything}            & ViT-L16   & MIX & Block2 & 51.4 & 31.6 & 28.4 & 42.2     & 30.2 & 23.5 & 16.2 & 6.8   & 95.2 & 68.1 & 35.1 & 17.5\\

DA~\cite{depthanything}            & ViT-L16   & MIX & Block3 & 58.9 & 48.6 & 49.7 & 53.5     & 29.8 & 21.4 & 16.8 & 9.3   & 90.9 & 67.8 & 47.9 & 30.5\\

DA(77K)~\cite{depthanything}              & ViT-L16   & MIX & Block0 &  8.0 &  5.8 &  5.2 &  6.7       & 18.3 & 15.1 & 10.7 &  5.0   & 63.6 & 34.9 & 20.4 & 12.4 \\

DA(77K)~\cite{depthanything}           & ViT-L16   & MIX & Block1 & 24.2 & 13.6 & 11.0 & 19.1       & 34.4 & 25.7 & 16.6 &  6.4   & 92.4 & 54.6 & 26.9 & 13.7 \\

DA(77K)~\cite{depthanything}           & ViT-L16   & MIX & Block2 & 50.8 & 31.0 & 28.0 & 41.6       & 43.4 & 32.9 & 23.2 &  8.9   & 94.9 & 67.0 & 34.9 & 17.8\\

DA(77K)~\cite{depthanything}            & ViT-L16   & MIX & Block3 & 53.6 & 42.2 & 43.4 & 47.7       & 38.4 & 29.8 & 21.7 & 11.8   & 92.8 & 71.9 & 50.7 & 31.0\\

Metric3Dv2~\cite{hu2024metric3dv2}              & ViT-L16   & MIX  & Block0 & 12.0 &  8.6 &  7.9 & 10.1        & 10.2 & 10.5 &  8.7 &  4.3  & 79.1 & 39.7 & 23.5 & 12.7 \\

Metric3Dv2~\cite{hu2024metric3dv2}       & ViT-L16   & MIX  & Block1 & 39.0 & 22.0 & 16.0 & 30.7        & 55.7 & 42.8 & 25.2 &  8.8  & 94.2 & 61.4 & 29.6 & 14.2 \\

Metric3Dv2~\cite{hu2024metric3dv2}                                       & ViT-L16   & MIX  & Block2 & 60.2 & 41.5 & 39.8 & 51.6        & 63.1 & 54.7 & 36.8 & 14.8  & 94.1 & 68.1 & 36.9 & 20.9 \\

Metric3Dv2~\cite{hu2024metric3dv2}                   & ViT-L16   & MIX  & Block3 & 53.6 & 42.3 & 42.8 & 48.0        & 59.5 & 50.3 & 35.1 & 16.3  & 86.6 & 56.5 & 29.6 & 17.2 \\

\midrule \rowheadergreen \multicolumn{16}{l}{\textit{\textbf{Generative Geometry Foundation Models}}} \\
Marigold~\cite{marigold}              & UNet   & MIX  & Block0 & 14.0 &  4.6 &  3.5 &  9.6      &  8.4 &  5.8 &  3.4 &  1.3   & 81.7 & 37.0 & 17.0 & 8.0 \\

Marigold~\cite{marigold}      & UNet   & MIX  & Block1 & 53.8 & 29.5 & 23.7 & 42.5      & 42.2 & 32.4 & 18.7 &  4.4   & 92.8 & 59.1 & 25.6 & 11.7 \\

Marigold~\cite{marigold}        & UNet   & MIX  & Block2 & 27.2 & 15.8 & 12.5 & 21.3      & 45.5 & 34.1 & 18.1 &  5.4   & 83.5 & 45.4 & 21.5 & 11.2 \\

Marigold~\cite{marigold}          & UNet   & MIX  & Block3 &  8.0 &  6.4 &  6.3 &  7.1      & 18.0 & 12.8 &  7.5 &  3.5   & 43.3 & 25.2 & 15.9 & 9.8 \\

DepthFM~\cite{depthfm}              & UNet   & MIX  & Block0 & 20.0 &  8.4 &  6.1 & 14.3    & 23.1 & 16.4 & 7.4 & 2.2    & 85.9 & 40.6 & 17.0 & 8.0 \\

DepthFM~\cite{depthfm}             & UNet   & MIX  & Block1 & 50.8 & 31.4 & 25.2 & 42.1    & 46.4 & 39.1 & 24.0 & 7.2    & 94.1 & 62.4 & 29.2 & 13.0 \\

DepthFM~\cite{depthfm}                                       & UNet   & MIX  & Block2 & 22.6 & 13.8 & 10.7 & 18.8    & 46.0 & 36.7 & 20.0 & 6.2    & 80.5 & 41.7 & 20.7 & 11.0 \\

DepthFM~\cite{depthfm}                                      & UNet   & MIX  & Block3 &  3.9 &  3.5 &  3.0 &  3.6    & 11.2 &  8.4 & 6.3 & 3.8   & 39.0 & 25.6 & 16.3 & 10.4 \\

GeowizardD~\cite{geowizard}              & UNet   & MIX  & Block0 & 13.7 &  4.7 &  2.9 &  9.7    & 8.0 & 5.1 & 3.2 & 1.34   & 81.9 & 35.1 & 16.6 & 8.5 \\

GeowizardD~\cite{geowizard}                    & UNet   & MIX  & Block1 & 41.3 & 19.1 & 13.4 & 31.2    & 43.0 & 32.5 & 16.9 & 3.8    & 89.3 & 52.3 & 22.5 & 10.7 \\

GeowizardD~\cite{geowizard}                              & UNet   & MIX  & Block2 & 20.2 & 11.4 &  8.4 & 16.3    & 38.3 & 27.1 & 12.8 & 3.7    & 71.1 & 35.4 & 17.8 & 10.1 \\

GeowizardD~\cite{geowizard}                             & UNet   & MIX  & Block3 &  8.5 &  5.7 &  5.8 &  7.2    & 13.8 &  9.9 &  5.4 & 2.7   & 32.5 & 20.1 & 12.7 & 8.1 \\

GeowizardN~\cite{geowizard}              & UNet   & MIX  & Block0 & 11.1 &  3.6 &  2.9 &  7.6   & 8.5  & 5.1  &  3.0 & 1.3    & 80.6 & 33.9 & 15.1 & 7.6 \\

GeowizardN~\cite{geowizard}                         & UNet   & MIX  & Block1 & 43.3 & 20.2 & 15.5 & 32.8   & 48.6 & 37.8 & 20.8 & 5.0    & 88.8 & 53.7 & 22.4 & 10.4 \\

GeowizardN~\cite{geowizard}                       & UNet   & MIX  & Block2 & 22.5 & 12.3 &  9.4 & 18.0   & 43.4 & 32.8 & 16.3 & 4.5    & 68.9 & 36.6 & 17.5 & 9.2 \\

GeowizardN~\cite{geowizard}                    & UNet   & MIX  & Block3 &  6.8 &  5.4 &  4.8 &  6.2   & 13.0 & 10.2 &  6.3 & 2.7    & 27.5 & 17.6 & 12.0 & 7.3 \\

GenPercept~\cite{genpercept}          & UNet   & MIX  & Block0 &21.5&   9.6&   7.0&  16.0   & 22.7& 16.1&  7.1&  1.8    & 84.4& 40.8& 17.3&  8.0   \\

GenPercept~\cite{genpercept}                              & UNet   & MIX  & Block1 & 62.0&  41.9&  34.4&  52.2    & 55.7& 46.4& 27.8&  6.4     & 94.5& 64.9& 29.7& 13.3   \\

GenPercept~\cite{genpercept}                          & UNet   & MIX  & Block2 & 28.2&  16.4&  13.3&  22.9   & 54.9& 43.0& 23.8&  6.1     & 84.5& 45.1& 21.5& 10.8  \\

GenPercept~\cite{genpercept}               & UNet   & MIX  & Block3 & 8.0&   5.9&   5.9&   7.0    & 26.6& 19.0& 10.6&  3.8     & 58.3& 31.4& 17.4& 10.0  \\
\bottomrule
\end{tabularx}
}
\end{table*}

\subsection{Surface Normal Estimation Datasets}

NYUv2 \cite{DATA_2012_NYUv2} is an real indoor dataset comprised RGB-D video sequences from a variety of indoor scenes captured from the Microsoft Kinect. We evaluate on the official test (654 images) set with the ground-truth surface normal generated by Ladicky \etal\  \cite{ladicky2014discriminatively}.

ScanNet \cite{DATA_2017_ScanNet} is a real RGB-D video dataset of indoor scenes. We use the ground-truth surface normal and test split (800 sampled images) provided by FrameNet \cite{huang2019framenet}. To mitigate the noise, it first computes two (X and Y) tangent principal directions by adopting the 4-RoSY field using QuadriFlow \cite{huang2018quadriflow} as proposed by TextureNet \cite{huang2019texturenet}, and the ground-truth normal can be directly computed as the cross product of them. 

DIODE \cite{DATA_2019_DIODE} 1024\x768 collects both outdoor and indoor scenes. It collects high-quality data, but it contains very low diversity with only 2 scenes for evaluation.

Sintel \cite{DATA_2012_SINTEL} is a synthetic dataset derived from an open-source 3D animated short film. We calculate the ground-truth surface normal with the provided ground-truth depth maps and intrinsic parameters following the depth-to-normal procedure of DSINE \cite{bae2024dsine}.

BEDLAM \cite{bedlam} contains synthetic monocular RGB videos with ground-truth 3D bodies with varying numbers of people in realistic scenes with varied lighting and camera motions. We calculate the ground-truth surface normal with the provided ground-truth depth maps and intrinsic parameters following the depth-to-normal procedure of DSINE \cite{bae2024dsine}.

Infinigen \cite{infinigen} generates diverse high-quality 3D synthetic scene data, which offers broad coverage of objects and scenes in the natural world with natural phenomena. The surface normal is rendered based on Blender.

MuSHRoom \cite{DATA_2024_MUSHROOM} is an indoor real-world multi-sensor hybrid room dataset, which contains 10 rooms captured by Kinect, iPhone, and Faro scanner. We use the ground-truth normal annotations supported by gaustudio \cite{ye2024gaustudio}.

Tank and Temples (T\&T) \cite{DATA_2017_Tanks_and_Temples} is a dataset including both outdoor scenes and indoor environments, whose ground-truth data is captured using an industrial laser scanner. We use the ground-truth normal annotations supported by gaustudio \cite{ye2024gaustudio}.

\begin{figure}
    \centering
    \includegraphics[width=1.0\linewidth]{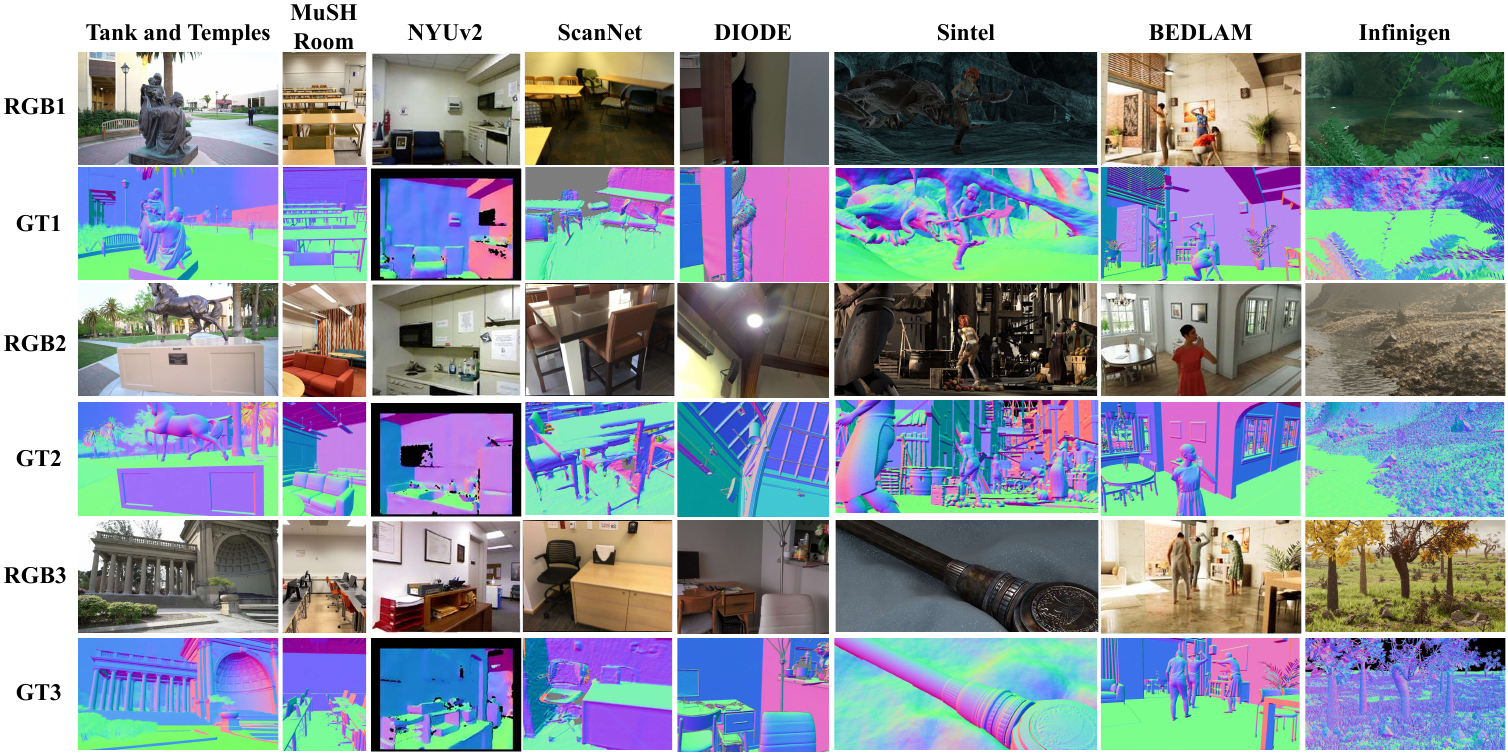}
    \caption{Visualization of the ground-truth surface normal from different datasets.}
    \label{fig:gt_norm_viz}
\end{figure}

\subsection{Limitations and Future Works}
The discussion of monocular depth estimation in this work is limited to single-image monocular affine-invariant depth estimation. Monocular metric depth estimation and video geometry estimation are also important topics, we leave them for future exploration.

\subsection{Broader Impacts}
In this section, we aim to discuss the potential societal impacts. The positive societal impacts encompass two aspects. First, it helps the research community gain in-depth knowledge about monocular geometry estimation, including performance comparisons between different models, technical details of current models, and future approaches. The release of this work also helps researchers perform experiments to evaluate their methods more comprehensively, fairly, and conveniently. Furthermore, it will significantly boost the progress of downstream tasks. As we mentioned in the paper, monocular geometry estimation can be applied to many downstream tasks, thereby accelerating their progress. In summary, we believe this work will have substantial positive effects on the research community, enriching the capacity of current and future applications and products, and ultimately improving people's lives. We also evaluated the negative societal impacts and found none.

\clearpage

{\bf Acknowledgements:}
This work was supported by National Key R\&D Program of China (No. 2022ZD0118700).

\bibliographystyle{alpha}
\bibliography{main}

\end{document}